\documentclass[pdflatex,sn-mathphys-num]{sn-jnl}

\usepackage{graphicx}%
\usepackage{multirow}%
\usepackage{amsmath,amssymb,amsfonts}%
\usepackage{amsthm}%
\usepackage{mathrsfs}%
\usepackage[title]{appendix}%
\usepackage{xcolor}%
\usepackage{textcomp}%
\usepackage{manyfoot}%
\usepackage{booktabs}%
\usepackage{algorithm}%
\usepackage{algorithmicx}%
\usepackage{algpseudocode}%
\usepackage{listings}%
\usepackage{hyperref}

\theoremstyle{thmstyleone}%
\theoremstyle{thmstyletwo}%

\theoremstyle{thmstylethree}%
\begin{document}

\title{Language Conditioned Multi-Finger Dexterous Manipulation Enabled by Physical Compliance and Switching of Controllers}


\author*[1]{\fnm{Cheng} \sur{Pan}}\email{cheng.pan@epfl.ch}

\author[1, 2]{\fnm{Kai} \sur{Junge}}\email{k.junge@embodiedai.ch}

\author[1]{\fnm{Benhui} \sur{Dai}}\email{benhui.dai@epfl.ch}
\author[1]{\fnm{Qinghua} \sur{Guan}}\email{qinghua.guan@epfl.ch}

\author*[1]{\fnm{Josie} \sur{Hughes}}\email{josie.hughes@epfl.ch}

\affil[1]{\orgdiv{CREATE Lab}, \orgname{STI}, \orgaddress{\street{Swiss Federal institute of Technology in Lausanne (EPFL)}, \city{Lausanne}, \country{Switzerland}}}

\affil[2]{\vspace*{1em}  \orgdiv{Embodied AI},  \city{Lausanne}, \country{Switzerland}}

\affil[]{\texttt{https://vla-dex-switch.github.io/}}




\abstract{
Human dexterity arises from the combined high-level task reasoning which is paired with finger-level dexterity control and physical compliance at the muscle and skin layers.
For robots, we see large Vision-Language-Action (VLA) models demonstrating text-conditioned high-level planning for a diverse range of manipulation tasks, typically performed with pincher grippers.
Smaller control policies developed through imitation learning are conversely showing success in more dexterous tasks using higher degree-of-freedom (DoF) grippers but for limited-scope tasks.
However, there are limited approaches for combining high-level cognition and reasoning with dexterous and robust low-level control, which requires both intelligent control and compliant robot design.
We propose a method inspired by the two-channel hypothesis of human motor control that combines these capabilities for dexterous manipulation, using a switching controller which integrates the best of high-level VLAs and smaller control models.
The coordination between these two channels is managed through an event-driven switching mechanism which monitors the subtask progression and completion, requiring only minimal demonstrations data by fine-tuning the VLA to predict event signals and training lightweight subtask-level dexterous policies.
This approach is applied to our custom compliant 13-DoF anthropomorphic dexterous robotic hand where we can modulate the compliance to systematically evaluate its impact on dexterity and robustness when combined with an autonomous policy.
With this, we show that only through the inclusion of hardware-level compliance in robotic fingers, which enables passive adaptation to disturbances and improves contact stability, can we mimic the robustness of human-like physical adaptation to tasks.
Specifically, the stability of finger-object contact is improved by over 38\% through the introduction of compliance.
The full methodology is validated across a range of language-conditioned dexterous tasks. To demonstrate the advantages of modularity, we show that the adaptation to additional dexterous skills and different compliant hands can be achieved without retraining the large VLA model.
This provides an efficient, scalable, and cross-embodiment approach to dexterity that actively leverages and exploits compliance whilst still leveraging the advantages of large AI models.

}

\maketitle

\section{Introduction}\label{sec1}

Achieving human-level manipulation and dexterity remains a central challenge of robotics, as it would enable robots to operate as versatile assistants in everyday environments \cite{billard2019trends, cui2021toward, ortenzi2019robotic}. Human manipulation capabilities arise from the ability to combine high-level task planning with low-level finger control capabilities for robustness and dexterous interactions with the environment \cite{yang2014cognitive}. 
At the high-level we show task understanding, object recognition, and reasoning about spatial relationships between the hand, objects, and the environment \cite{andersen2009intention}. 
However, to physically interact with objects we require dexterous low-level control at the finger level, allowing the hand to perform precise finger–object interactions during sequential manipulation subtasks \cite{sobinov2021neural}.
In humans, this is further combined with physical intelligence at the contact level, through compliance and variable stiffness in the hand and fingers, enabling adaptation to unpredictable environmental disturbances and motor control errors \cite{burdet2001central}.
Humans dynamically coordinate this hierarchical control and distributed intelligence during manipulation. At the beginning of a task, attention is directed toward high-level planning, but as the hand approaches the object, attention shifts toward fine-grained control, focusing on finger-level control and environmental interactions.  When disturbances or uncertainties occur, our bodies physically react, providing contact level robustness and resilience. 
In robotics, individual components of this hierarchical control have been demonstrated in isolation, however, integrating these abilities into a unified system capable of performing diverse manipulation tasks for anthropomorphic hands remains challenging \cite{billard2019trends}. 

High-level planning for manipulation is rapidly advancing with the development of generalist manipulation models. 
Vision–Language–Action (VLA) models demonstrate impressive capabilities across diverse manipulation tasks \cite{kawaharazuka2025vision, shao2025large}, leveraging pretrained vision-language backbones and training on extensive datasets, including internet-scale visual data and robotic trajectories \cite{zitkovich2023rt, black2024pi_0, intelligence2025pi_}. VLA models show proficiency in high-level reasoning, language understanding, target object recognition, and general task adaptation \cite{kawaharazuka2025vision, shao2025large, zhao2025cot}. These strengths enable robust performance across a wide range of environments and diverse manipulation tasks.
Despite these advantages, VLA models are primarily limited to coarse manipulations, such as pick-and-place, object transport, and cloth folding, where most motions involve arm-level trajectories rather than fine gripper control \cite{sapkota2025vision, kim2024openvla, black2024pi_0}. Additionally, the majority of VLA pretraining data is collected using simple 1-degree-of-freedom (DoF) parallel grippers \cite{o2024open, khazatsky2024droid}, limiting application to high-DoF manipulators such as anthropomorphic robot hands which can demonstrate more dexterity. Adapting these large models to such settings typically requires expensive repeated large dataset collection and end-to-end retraining, which is computationally expensive and difficult to scale. Furthermore, refining VLA performance to specific subtasks often requires retraining or fine-tuning the full model, which is computationally expensive and pose a risk of catastrophic forgetting or reduced performance on other tasks \cite{kawaharazuka2025vision, sapkota2025vision}.

For more fine-grained dexterous manipulation which instead focuses on precise control transformer-based \cite{zhao2023learning} and diffusion-based policies \cite{chi2025diffusion} have demonstrated success, showing deformable object manipulation \cite{ze20243d, zhao2024aloha}, high-precision insertion and assembly \cite{wu2025tacdiffusion}, and contact aware surface interactions \cite{chi2025diffusion, pan2025online}. These models can also be adapted for manipulators with a high degree of freedom, such as multi-fingered hands, enabling dexterous tool use \cite{ze20243d, xu2025dexumi} and in-hand object manipulation tasks \cite{qi2025simple}. Compared to VLA, these small policy models require significantly less training data and can be tailored to specific tasks, achieving precise and robust control at the finger level. However, they are typically restricted to specific motions or task types. Their lack of extensive pre-trained knowledge also limits their ability to generalize across diverse tasks.

At the physical interaction level, the impact of the robot's mechanical design, specifically its compliance, has a dominant effect on its behavioral capabilities and robustness \cite{piazza2019century, gilday2023sensing, frost1994perspectives, hughes2018anthropomorphic}.
Without even sensory feedback and control, soft robotic fingers and grippers such as the PISA-IIT soft hand\cite{santina_toward_2018} and Yale hand\cite{Ma_underactuated_2014} show excellent adaptation to target objects. 
Dexterous tasks have also been studied with physically compliant hardware such as the RBO Hand 3\cite{deimel2013compliant}, where open-loop pneumatically actuated hands are able to perform in-hand cube rotations robustly, adaptive to a variety of object geometries \cite{patidar2023hand}. 
More broadly, compliance in the manipulator arm has repetitively been shown to assist contact with unknown surfaces or environments \cite{wu2021learning, junge2025adapt} .
While the quantification of the robustness through compliant robotic hands is explored under simple control algorithms\cite{junge2025spatially}, its use with learning based controllers is unexplored.

\begin{figure*}[tb]
     \centering
     \includegraphics[width=1.0\textwidth]{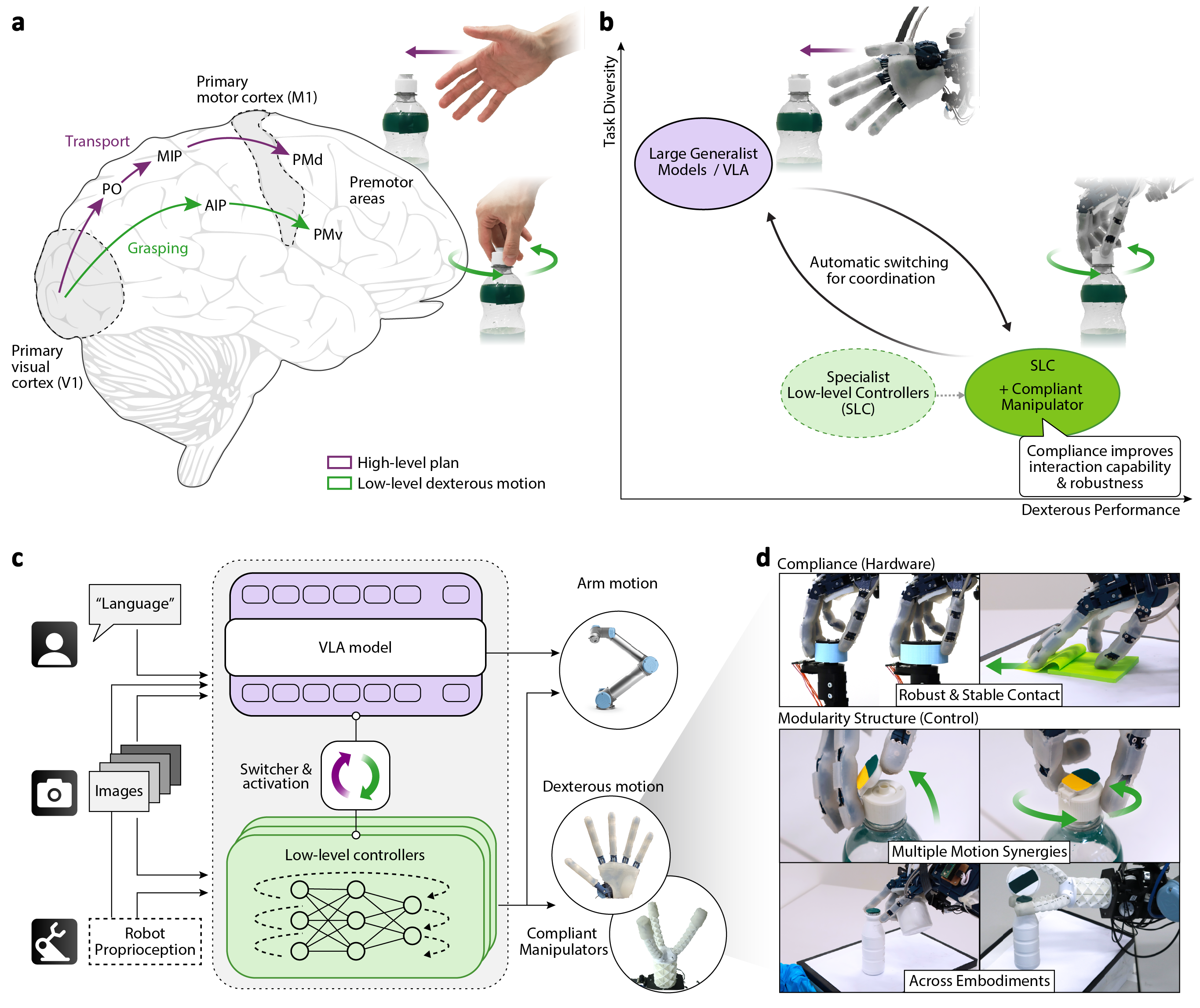}
     \caption{Switching-based dual-channel control for dexterous manipulation. \textbf{a}, Two-channel hypothesis in neuroscience: human manipulation motion commands are generated by two neural pathways. The reaching channel (purple) governs arm transport, while the grasping channel (green) enables dexterous motion. \textbf{b}, Combining the complementary strengths of large generalist models (VLA) and specialized low-level dexterous controllers, enhanced robustness through compliance. \textbf{c}, Bio-inspired dual-channel control pipeline, an event signal is used to switch between models. \textbf{d}, Demonstration of language-conditioned dexterous skills. PO, parieto-occipital area; MIP, medial intraparietal area; PMd, dorsal premotor cortex; AIP, anterior intraparietal area; PMv, ventral premotor cortex;
     }
    \label{fig1}
\end{figure*}

To advance dexterous manipulation, we must establish how to integrate the complementary capabilities of VLAs, which offer general purpose visual-language reasoning, with that of specialized dexterous control policies such as diffusion policies, which provide fine-grained dexterity.
To be usable, such integration must be data and computation efficient: it should avoid large-scale data collection or retraining of large VLA model from scratch when adapting to new tasks or different dexterous manipulators.
We must also understand how to leverage physical adaption and intelligence through the introduction of compliance, and how this can compensate or augment the robustness of low-level behaviors. 
Insights from human motor control, including the two-channel hypothesis \cite{jeannerod1999visuomotor}  (Fig. \ref{fig1} a), provide a useful perspective on the integration challenge. 
This model predicts that distinct neural pathways link visual cortex and motor cortex for manipulation \cite{whishaw2016dissociation}. The transport-channel guides the arm toward a target using extrinsic spatial information (e.g., object position), while the grasping-channel governs the shaping and coordination of the fingers based on intrinsic object properties (e.g., size and shape). Successful manipulation arises from the temporal and spatial coordination of these two channels, enabling simultaneous arm transport and finger control.

Motivated by this biological principle, we propose a switching-based dual-channel architecture combining the complementary strengths of large generalist models and specialized dexterous controllers.
This uses a switching method that is able to move between the large VLA model, and a low-level controller, enabling dexterous completion of a large range of tasks (Fig.\ref{fig1}c)
The VLA model serves as the high-level arm transport channel, responsible for task understanding and language-conditioned planning, generating arm motion that brings the hand to the target object. A dexterous low-level controller serves as the grasp channel, generating precise finger-level motions required for dexterous manipulation subtasks. 
The coordination between these two models is realized through a switching mechanism driven by an event signal, predicted from both channels (see Fig. \ref{fig1}c). The event signal indicates the completion of a subtask and triggers the transition of control between the VLA planner and the dexterous low-level controller.
In addition to algorithmic design, we further hypothesize that hardware-level compliance can improve robustness during physical interaction. Human manipulation is inherently tolerant to small errors and disturbances. Rather than maintaining perfectly precise positions or forces, humans continuously compensating for small errors and disturbances through compliance. Inspired by this property, we introduce a custom, anthropomorphic robotic hand with series elastic elements at the base joints of the robotic fingers building on the ADAPT hand~\cite{junge2025adapt}. This provides passive compliance that allows the hand to absorb disturbances and maintain stable contact during manipulation.

In this paper we show how this switching-based dual-channel architecture enables the integration of world knowledge of the VLA models with dexterous capabilities. Crucially, our approach is  data-efficient: instead of retraining VLA models on large-scale dexterous datasets, we require only a small amount of data to (i) fine-tune the VLA to predict event signals and (ii) train lightweight dexterous policies at the subtask level. This allows pretrained VLA models to be used for high-Dof anthropomorphic robot hands. Dexterous low-level controllers act as a flexible interface that adapts the high-level policy to different manipulation subtasks and manipulators (see Fig. \ref{fig1}d). 
Furthermore, the compliant mechanical design of the robotic hand complements this framework by enabling passive adaptation to environmental disturbances; the compliance improves contact robustness during manipulation by 38\%.
We further demonstrate the effectiveness of the switching-based framework on a range of tasks, including language-conditioned pick-and-place, water bottle lid opening and pouring, sticky note peeling and pasting, and multi-object grasping and disposal.
The modular design of dual-channel framework offers several additional practical benefits. Low-level controllers can be independently updated to improve specific manipulation behaviors or adapt to different manipulators, sensing modalities or motion synergies (see Fig. \ref{fig1}d), without retraining the large VLA model. This separation reduces both data requirements and computational cost while preserving the general capabilities of the pretrained model.
Together, the switching-based dual-channel control structure and physical compliance allow the robot to achieve general, dexterous and robust manipulation performance.

\section{Results}\label{sec2}

In this section, we first demonstrate the compliant robot hand and teleoperation system for high-quality data collection, highlighting how compliance improves contact robustness and dexterous performance. We then present the autonomous control capabilities of each component—dexterous low-level controllers and the VLA model—within a switching-based dual-channel architecture coordinated by an event signal. These complementary models are integrated into a modular pipeline for dexterous manipulation with an anthropomorphic hand across varying tasks. The framework’s modularity enables efficient adaptation to new motions and manipulators, as well as targeted subtask performance improvements through component updates.


\begin{figure*}[tb]
     \centering
     \includegraphics[width=1.0\textwidth]{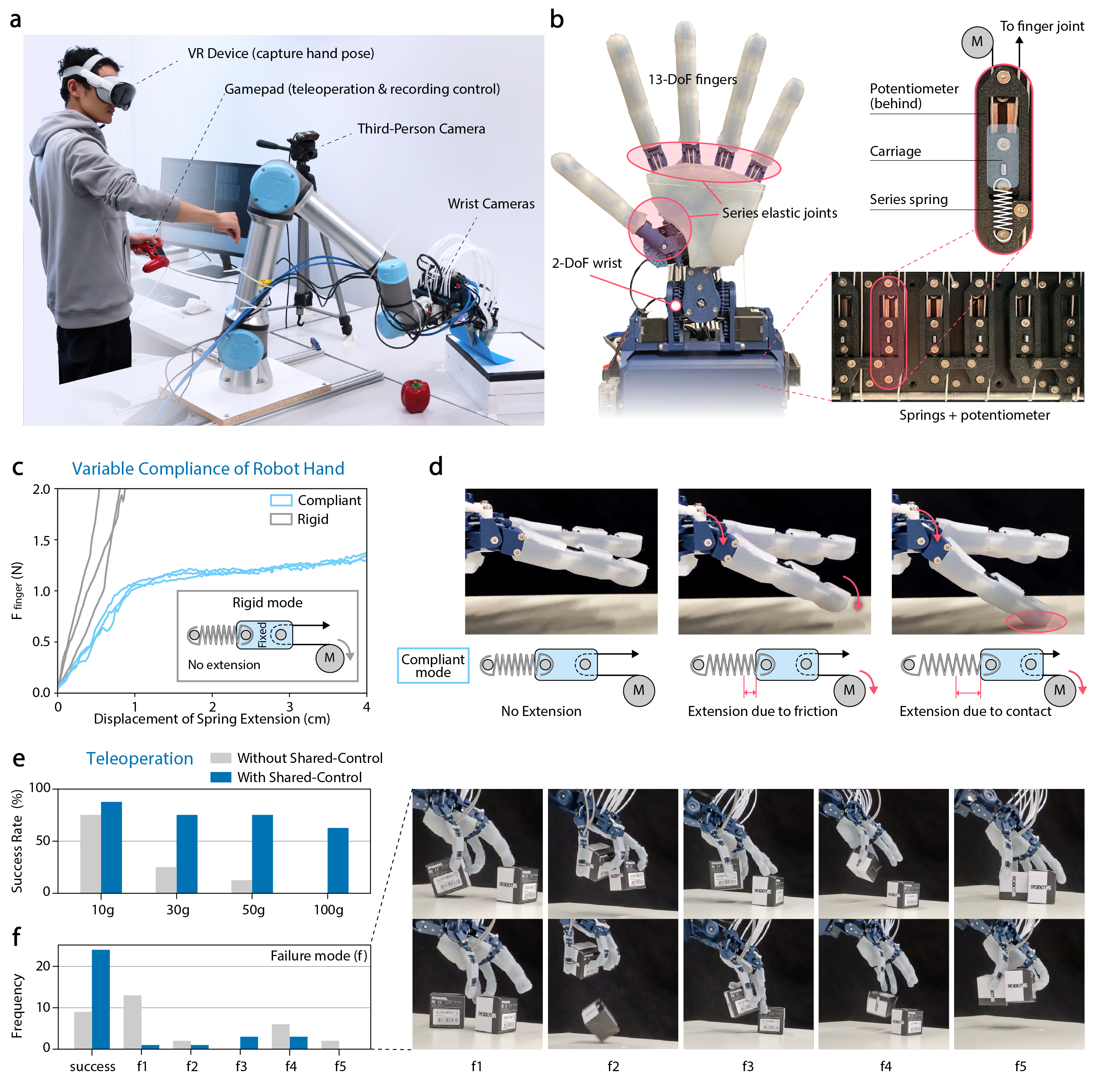}
     \caption{Teleoperation of the anthropomorphic robot hand with variable compliance. a, Teleoperation hardware. b, the 15-DoF anthropomorphic robot hand with variable compliance, Adapt Hand \cite{junge2025adapt}. c, Force–Displacement characterization of the ``compliant'' and ``rigid'' configurations, including mechanism of the rigid mode. d, Compliant mode realized through extension of the series spring, e-f, Shared-control teleoperation experiment results: success rate and frequency of failure modes in the multi-object grasping task. Failure modes: f1, Cannot lift either of boxes; f2, Dropped after lifting both boxes; f3, Positioning of first box prevents grasping of second box; f4, Drops first box in mid-air; f5, Momentarily drops the box(es) and regrasps.
     }
    \label{fig:hardwares}
\end{figure*}

\subsection{Compliant Robot Hand and Teleoperation}

\subsubsection{Robot Hand with Variable Compliance}

Compliance can improve manipulation robustness in contact-rich tasks and reduce reliance on complex haptic controllers or highly precise motion-mapping schemes.
To demonstrate the role of compliance, we have developed a robot hand where we can vary the compliance, an adaption of the ADAPT Hand (ADH) \cite{junge2025adapt} (Fig. \ref{fig:hardwares}b and Supplementary Fig. \ref{si_fig:ADH}). 
The hand provides 13 degrees of freedom (DoF) across the fingers and an additional 2-DoF at the wrist. It incorporates series elastic elements at the finger/thumb base joints, the metacarpophalangeal (MCP) and carpometacarpal (CMC) joints, to introduce mechanical compliance that enables adaptive finger motion.  
Compliance is realized through series springs that modulate tendon tension in response to external interaction forces. The extension of each series spring is measured by a potentiometer and serves as an indirect estimate of the forces experienced at the corresponding base joint. Fig. \ref{fig:hardwares}d illustrates spring extension arising from friction and object contact during operation in the ``compliant" mode.

As shown in Fig. \ref{fig:hardwares}c, the ADH can also be modified to operate in a ``rigid" mode by mechanically constraining the seires elastic elements. Fig. \ref{fig:hardwares}c compares the force–displacement characteristics of the compliant and rigid configurations. While the rigid mode exhibits an approximately linear force–displacement relationship, the compliant mode produces a plateauing force profile that smooths the increase in fingertip force and enables near-constant force over large displacements. This tunable transition between compliant and rigid configurations enables systematic evaluation of their respective effects on autonomous manipulation performance.

\subsubsection{Teleoperation with Shared-Control}

To capture high-quality demonstration data with this robot hand, we require intuitive teleoperation across both the rigid and compliant modes. 
To do this, we use a virual reality (VR) device (Apple Vision Pro) to track the teleoperator’s hand and wrist motions (Fig. \ref{fig:hardwares}a) which are then directly mapped to control commands for the robot hand and arm. During demonstrations, both robot and environmental states are recorded. These states include control commands, the actual joint positions of the robot arm and hand as proprioceptive states, and visual observations from a third-person camera and wrist-mounted cameras on the robot hand. These demonstrations serve as the training data for learning autonomous control models.

Although teleoperation allows control of high-DoF anthropomorphic robot hands, demonstrating dexterous manipulation remains challenging and often generates noisy demonstrations. 
This results from embodiment mismatches between human and robot hands, VR tracking errors from occluded finger motions, and absence of haptic feedback, which forces operators to visually infer contact and inter-finger opposition states.
We propose to increase the capabilities of teleoperation and the quality of training data through shared-control, which leverages the proprioceptive springs in the hand. 

The shared-control framework allows the robot to be teleoperated while maintaining a reliable force level with the object in contact autonomously. 
Finger contact and alignment states are estimated from potentiometer readings in the series elastic elements (Figs. \ref{fig:hardwares}b and \ref{fig:hardwares}d) and are used as inputs to the shared controller to modulate motor commands for the robot hand.
Based on these readings, the shared controller assists the operator's intent to apply force on the object, higher than what could be achieved via pure teleoperation. 
Through a state machine, the shared controller is able to switch back and forth between a purely teleoperated state and a contact-assisted state automatically, such that the operator does not need to manually switch between the two.

We evaluate the benefits of the proposed shared-control method using a multi-object grasping task. This task requires coordinated finger motion robust and contact control.
As shown in Fig. \ref{fig:hardwares}e, the shared-control framework enables reliable teleoperated multi-object grasping of two boxes with individual weights ranging from 10 to 100 g, yielding a substantial improvement in success rate compared to direct teleoperation. 
At the lowest object weight, both systems (without and with shared-control) achieve high success rates, 75\% and 87.5\% respectively. 
As the object weight increases, requiring more stable force to maintain contact and grasp, performance degrades for both approaches. However, the shared-control system maintains robust performance. Notably, when grasping two boxes weighing 100 g each, the success rate with shared control remains at 62.5\%, whereas the success rate without shared control drops to zero. These results demonstrate that shared control significantly improves teleoperation robustness and reliability, particularly under increased force and coordination demands, thereby enabling more consistent collection of high-quality demonstration data.

Failure modes observed during the multi-object grasping task and their frequencies are summarized in Fig. \ref{fig:hardwares}f. These failures highlight the difficulty of maintaining stable object contact and correct inter-finger positioning during direct teleoperation, especially in the absence of shared control. By contrast, the shared-control framework substantially reduces the frequency of teleoperation failures while also lowering the control burden on the human operator. Moreover, the shared controller filters noise in the hand motion data, arising from occlusion of finger movements by the palm and consequent inaccuracies in VR-based hand tracking, resulting in smoother trajectories and improved demonstration quality.


\subsection{Compliance-Driven Robustness and Dexterity of Low-level Controller}

\subsubsection{Compliance Enhances Robustness in Contact-Rich Manipulation}

\begin{figure*}[tb]
     \centering
     \includegraphics[width=0.89\textwidth]{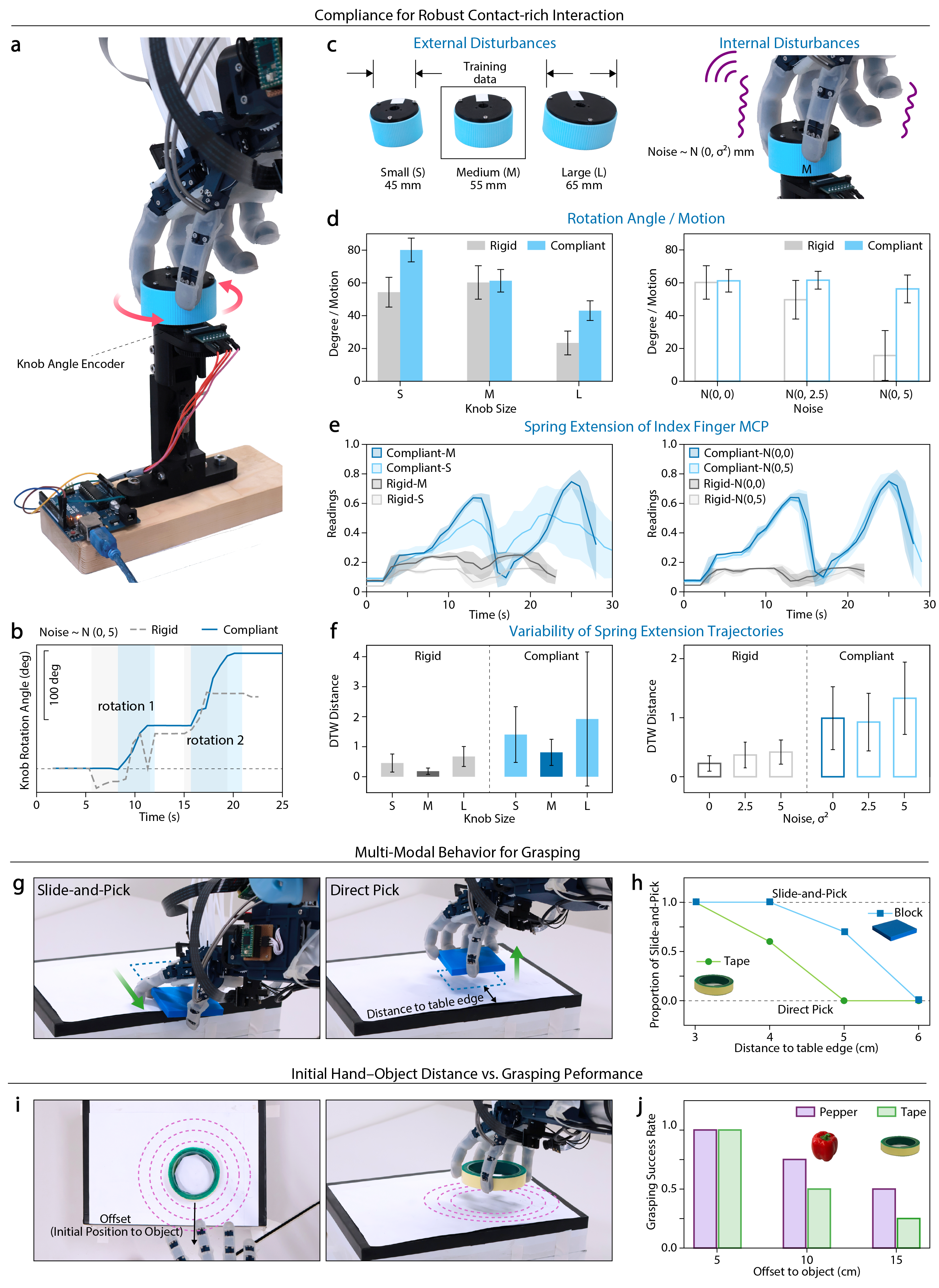}
     \caption{Robustness (enhanced by compliance) and performance of dexterous low-level controller. a, Knob-turning task for quantitatively evaluating robustness of dexterous low-level controller (Supplementary Video 1). b, Example trajectories of knob rotation angle using the robotic hand with rigid and compliant configurations, under simulated control noise N(0, 5)mm. c \& d, Results about rotated angle per turning motion with external disturbances (object size variance) and internal disturbances (control noises). e \& f, Measured spring extension of finger, indicating adaptive behavior compliant robot hand. g, Multi-modal behavior of dexterous low-level controller, picking up flat objects using two motion synergies. Probability distribution of using slide-and-pick synergy across the distance of object to the table edge (Supplementary Video 2). h, Performance sensitivity to uncertain starting configuration (offset to the object) of the dexterous low-level controller.
     }
    \label{fig:low-level_results}
\end{figure*}

To assess the benefits of compliance during dexterous autonomous manipulation in contact-rich scenarios, we consider a knob-turning task (Fig. \ref{fig:low-level_results}a). This requires coordinated finger motions and precise control of contact forces between the fingers and the knob surface. Manipulation performance is quantified by the rotation angle of the knob (Fig. \ref{fig:low-level_results}b), measured using an integrated encoder.

We compare robustness to disturbances between the rigid and compliant configurations of the robot hand, illustrated in Fig. \ref{fig:hardwares}b–\ref{fig:hardwares}d. For each configuration, a diffusion policy \cite{chi2025diffusion} is trained as the low-level controller using demonstrations collected on a medium-sized knob (Fig. \ref{fig:low-level_results}c). Task performance is quantified as the average knob rotation angle per motion, computed after two consecutive autonomous turning motions.
To evaluate robustness under disturbances, we introduce disturbances along two dimensions, external and internal (Fig. \ref{fig:low-level_results}c). First, external uncertainty from environment is introduced by varying the knob size during testing. Second, internal control errors are simulated by injecting Gaussian noise into the robot arm’s planar (x-y) position commands. The Gaussian noise is defined as $N(0,\sigma^2)$, where 0 is the mean and $\sigma$ the standard deviation. For each disturbance condition, ten trials are conducted for both rigid and compliant configurations using their respective trained diffusion controllers.

Results in Figs. \ref{fig:low-level_results}d–\ref{fig:low-level_results}f and example knob angle trajectories in (Fig. \ref{fig:low-level_results}b) show that the compliant hand achieves more effective knob rotation than the rigid configuration under both external disturbances (knob size variation) and internal control errors (simulated via Gaussian noise added to the control output), as it provides more stable finger–object contact. When the knob diameter is varied, the compliant configuration outperforms the rigid configuration by an average of 58\%. For the compliant hand, smaller knobs yield larger rotation angles than medium and large knobs. This trend arises because each turning motion corresponds to a fixed arc-length displacement induced by finger motion, with the resulting rotation angle inversely proportional to the knob diameter. Notably, although performance degrades for the rigid configuration as knob size deviates from the training condition, the compliant hand maintains robust performance under external disturbances, reflecting its superior adaptive capability. 
A similar pattern is observed under internal control disturbances. The compliant configuration exhibits less than an 8\% performance decrease in the presence of control noise compared to the noiseless case, whereas the rigid configuration suffers a substantial performance drop of 74\% when Gaussian noise N(0,5)mm applied. In the rigid configuration, which lacks automatic adjustment of force and contact, disturbances can lead to either insufficient or excessive contact forces. Too little force fails to trigger interaction with the target object, whereas excessive force can displace the object into an incorrect state, resulting in failures or reduced manipulation performance. In contrast, mechanical compliance enhances robustness to both environmental variability and control errors by stabilizing contact interactions and automatically adjusting applied forces in response to external or internal perturbations. This adaptive behavior leads to more reliable and consistent autonomous manipulation performance.

\subsubsection{Adaptive Behavior Enabled by Compliance}
The adaptive capability of the compliant hand is further illustrated by the spring readings (Figs. \ref{fig:low-level_results}e–\ref{fig:low-level_results}f). The series elastic elements compensate for disturbances arising from both knob size variations and control noise by adaptively adjusting tendon extension to maintain stable contact and force at the fingers. Fig. \ref{fig:low-level_results}e shows the MCP joint spring readings of the index finger, with the mean across ten trials indicated by the solid line and the variability represented by the shaded area. The results demonstrate increased variance in spring extension in the presence of disturbances, particularly for varying knob sizes. Spring readings of all fingers are illustrated in Supplementary Fig. \ref{si_fig:spring_size}-\ref{si_fig:spring_noise}.

To quantify the variability of spring readings, we compute the Dynamic Time Warping (DTW) distance between spring trajectories. This metric captures intrinsic differences across trajectories while reducing the effect of temporal misalignment, since finger motions may occur at slightly different timepoints or rates across trials. The observed increase in DTW distance under both disturbance conditions confirms larger spring fluctuations, reflecting the hand’s adaptative response to external and internal perturbations. These results highlight the role of mechanical compliance in maintaining stable finger-object contact and robust manipulation performance, providing a quantitative measure of the adaptive behavior that underlies disturbance resilience.

\subsubsection{Learning Multi-model Behaviors from Demonstrations}

High-DoF manipulators allow for diverse manipulation strategies, enabling multiple valid solutions to the same task. Human experts naturally demonstrate such multimodal behaviors, providing rich supervision for policy learning. To evaluate whether diffusion policies can capture not only dexterous finger motions but also multimodal manipulation strategies that exploit environmental affordances, we design a grasping task involving two thin and flat objects, a blue block and a tape (Fig. \ref{fig:low-level_results}g).

In this task, the object can be grasped using two distinct strategies. When the object is located away from the table edge, a ``Direct Pick" strategy is used, requiring precise vertical alignment of all fingers with the object to achieve a stable grasp. When the object is placed near the table edge, an alternative ``Slide-and-Pick" strategy is employed: the object is first slid toward the edge, enabling the thumb to engage the object from below and simplifying the grasp. A diffusion-based policy is trained using human demonstrations that include both strategies, with an equal number of trajectories for each grasp type.

During evaluation, the object is placed at varying distances from the table edge, and grasping behavior is assessed over five trials per location. As shown in Fig. \ref{fig:low-level_results}g, the probability of selecting the slide-and-pick strategy increases as the object approaches the table edge, while the direct pick strategy dominates when the object is located farther from the edge. This behavior mirrors human manipulation strategies that leverage environmental context to simplify task execution. These results indicate that the low-level diffusion policy successfully learns multimodal manipulation behaviors and adapts its strategy based on environmental conditions. Moreover, all grasping trials succeed across all object placements, demonstrating the effectiveness of the learned policy in executing challenging grasps with the high-DoF robotic hand.

\subsubsection{Performance Sensitivity to Uncertain Initial Configuration}
To ensure reliable dexterous motion across diverse initial configurations, it is essential to evaluate the limitations of the dexterous low-level controller.
Previous experiments demonstrate that the low-level diffusion policy can precisely control the robotic hand and achieve high grasping success rates, but only when the hand begins sufficiently close to the target object, with the valid region determined by the spatial coverage of the training demonstrations. 
In our dataset, demonstrations typically begin within approximately 5 cm of the target object; larger offsets therefore constitute out-of-distribution initial conditions and can degrade performance.

To characterize this limitation and motivate the need for a general high-level planner (VLA), we systematically offset the hand’s initial position relative to the object in the planar x-y workspace parallel to the tabletop and evaluate grasping performance from these starting configurations. The evaluated starting regions are illustrated in Fig. \ref{fig:low-level_results}g. Grasping experiments are conducted for two objects (pepper and tape), with eight trials performed at each offset distance. When the initial hand-object offset exceeds 15 cm, grasping success rates drop below 50\% for both objects. These results highlight the sensitivity of the low-level controller to initial conditions outside the training distribution and underscore the necessity of a high-level planner to reposition the robot hand into a suitable initial region before performing dexterous manipulation. The results about how closely the high-level VLA model can bring the manipulator to the target object are shown in Supplementary Fig. \ref{si_fig:vla_dist}.


\begin{figure*}[tb]
     \centering
     \includegraphics[width=1.0\textwidth]{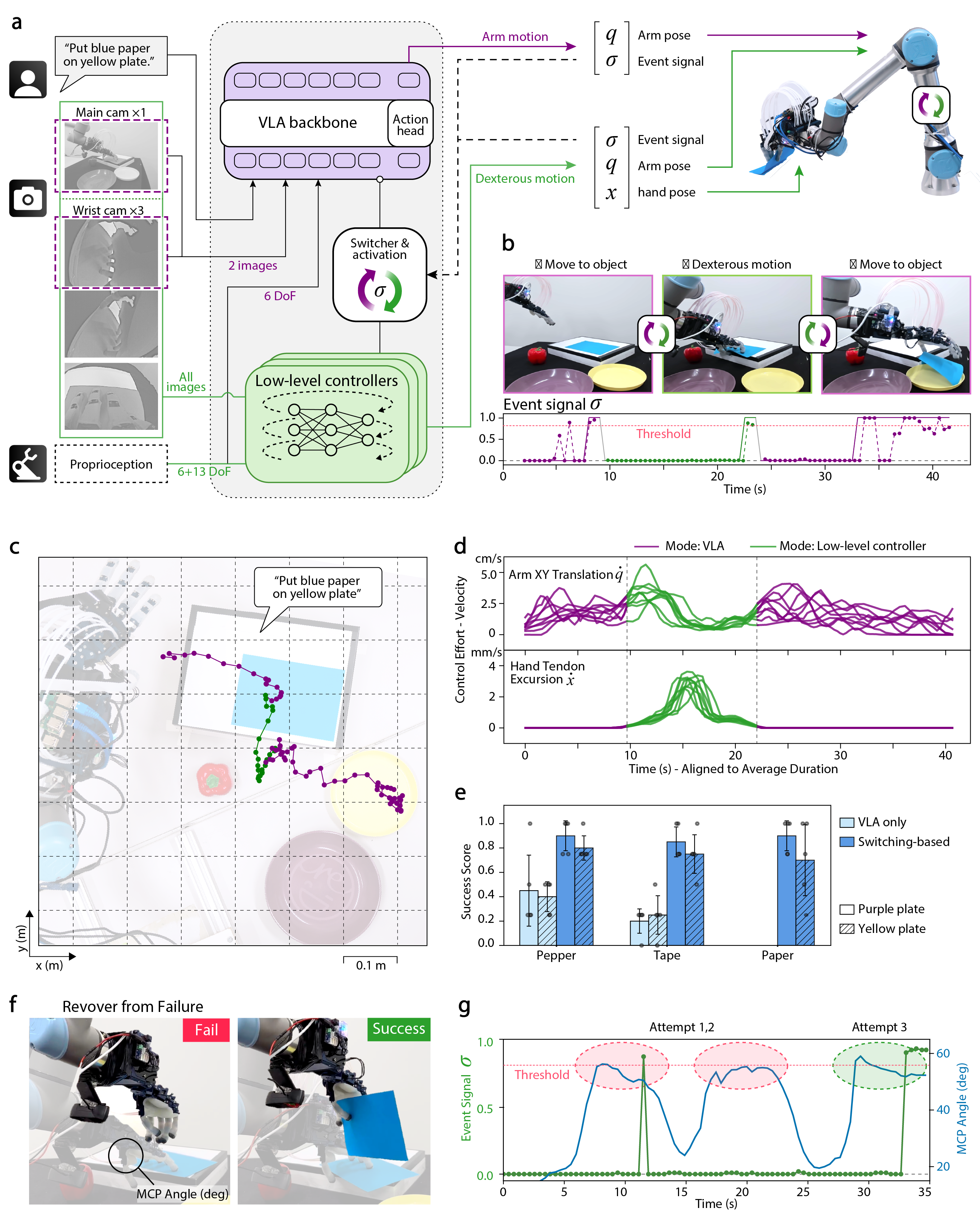}
     \caption{Coordination of VLA and Dexterous Low-Level Controllers. a, Switching-based dual-channel control framework. b, Pick-and-place example task and the event signal for switching between two channels (Supplementary Video 3). c, Example planar trajectory. d, Control effort of coordinated models. e, Success rate of the task under different language instructions. f-g, Recovery behavior from failed grasping and the event signal response with actuator (motor) angle of the thumb MCP joint (Supplementary Video 4).}
    \label{fig:switch_method}
\end{figure*}

\subsection{Dual-Channel Architecture and Switching-based Coordination}
\label{sec:coordinate}
To leverage both the generalization capabilities of a pretrained vision–language–action (VLA) model and the dexterous control provided by a low-level diffusion policy, we adopt a dual control channel architecture with an event-driven switching mechanism. Given a natural language command, the VLA model guides the robot arm and hand toward the target object. Once a suitable pre-manipulation configuration is reached, control is transferred to the low-level controller to execute dexterous manipulation. Upon completion of the subtask, control is returned to the VLA model to proceed with subsequent actions. 
Analysis about the control effort profiles shows
the complementary roles of the VLA and low-level controllers.
We also demonstrate that the event signal for switching is robust to grasping failures and recovery attempts, and reliably predicts true sub-task completion. 

\subsubsection{Dual-Channel Control Framework}
 The VLA model and the low-level controller alternately control the robot arm and hand through an event-driven switching mechanism. The overall framework is illustrated in Fig. \ref{fig:switch_method}a. The VLA model serves as a high-level planner, taking as input a natural language command, visual observations from a third-person and camera a wrist-mounted camera, and the 6-DoF proprioceptive state of the robot arm. Based on the language instruction and sensory inputs, the VLA model generates arm-level action commands $q$ to move the robot hand toward the target object. In addition to action commands, the VLA model predicts an event signal $\sigma$ that governs the control switch. This event signal remains near zero while the VLA model is actively guiding the robot. Once the hand reaches an area close to the target object-indicating completion of the high-level subtask, $\sigma$ transitions to one, triggering the activation of the low-level controller.

The dexterous low-level controller, implemented as a diffusion-based policy, receives “higher-resolution” inputs to enable precise dexterous control. These inputs include one third-person view image, three wrist camera images, and a 19-DoF proprioceptive state vector encompassing both the robot arm and hand. The controller outputs commands for both arm pose refinement $q$ and dexterous finger motions $x$. Similar to the VLA model, the low-level controller also predicts $\sigma$. When the dexterous manipulation subtask is completed, $\sigma$ transitions to one from zero, returning control to the VLA model for subsequent task execution.

Through this event-based coordination between the high-level VLA planner and the low-level diffusion policy, the proposed switching-based framework enables both general language-conditioned task execution and precise dexterous manipulation with a high-DoF robotic system.

\subsubsection{Event-Driven Channel Switching}
An event signal $\sigma$ is introduced to coordinate control between  the high-level VLA model and the low-level controller. As illustrated in Fig. \ref{fig:switch_method}a, the switcher module uses $\sigma$ as input to determine which controller is active, enabling or disabling inference and command execution from either the VLA or the low-level controller.
The event signal represents a binary indicator of subtask completion. Both the VLA model and the low-level controller predict $\sigma$ during execution. When one controller is active, its predicted $\sigma$ is continuously monitored. If $\sigma$ satisfies the switching condition - indicating completion of the current subtask, control will be transferred to the other controller to execute the next subtask.


An example of event-driven switching is shown in Fig. \ref{fig:switch_method}b for a pick-and-place task. 
The event signals are visualized in the lower panel of Fig. \ref{fig:switch_method}b, where solid lines denote ideal ground-truth $\sigma$ and dashed lines indicate the predicted $\sigma$ from models. 
Upon receiving a natural language instruction, the VLA model initially controls the robot arm to move the hand toward the first target object (“blue paper”). During this phase, the VLA is active and its $\sigma$ (purple) remains near zero and transitions toward one as the hand approaches the object. For the switching mechanism, a switching threshold is applied, and control transitions occur only when the $\sigma$ exceeds a predefined threshold for a fixed number of consecutive time steps (switching condition) to ensure robustness against transient prediction noise.
Once $\sigma$ satisfies the switching condition, control is transferred to the channel of low-level diffusion policy, which executes dexterous finger motions to grasp the object. Similarly, the low-level controller’s $\sigma$ (green) remains near zero during grasp execution and transitions to one upon successful grasp completion.
Control is then returned to the VLA model when the switching condition is filled, which fixes the hand pose to maintain a stable grasp and moves the object toward the second target (“yellow plate”). When the hand approaches the placement location, the VLA's $\sigma$ again transitions to one, triggering a hand opening action to release the object. This open-loop releasing motion is specifically designed to replace low-level controller for this step considering the simplify of releasing motion. In this particular task, execution terminates after this second VLA's event-driven switching; however, the same mechanism can return control to the low-level controller for longer-horizon manipulation sequences (see Sec. \ref{sec:dexDemos}).

To enable reliable event prediction, the VLA model is finetuned using demonstration data with event signal supervision. This event signal design closely aligns with an existing binary control channel in the pretrained VLA model originally used to regulate gripper openness. In the pretrained setting, this channel typically transitions when the gripper approaches or contacts an object and remains unchanged otherwise. During finetuning, this channel is repurposed to represent subtask completion, with the target event signal similarly transitioning to one when the hand reaches the target object and zero otherwise. This semantic and structural similarity facilitates effective knowledge transfer during finetuning, allowing the VLA model to robustly predict event signals with minimal retraining.

\subsubsection{Control Effort Analysis of Coordinated Models}
To illustrate the distinct roles of the high-level VLA model and the low-level controller, we visualize one representative execution trajectory in Fig. \ref{fig:switch_method}c, with segments color-coded by the active controller. From a top-down view, the trajectory generated by the VLA model (purple) spans a large region in the planar x-y workspace, whereas the trajectory generated by the low-level controller (green) is localized around the target object. This spatial separation reflects the functional division between the two models: the high-level VLA model primarily governs arm-level motion for object approach and relocation, while the low-level controller focuses on dexterous finger manipulation near the object.

We further quantify the control effort in terms of velocities of commands, in order to highlight the distinct control behaviors of the VLA and low-level policies. The control effort is computed as step-wise differences in arm pose  $\dot q$ and hand joint angles $\dot x$ produced by each controller, averaged across all action dimensions.
Control effort is evaluated separately for planar arm motion and for finger joint actuation (more results about arm vertical translation and orientation are shown in Supplementary Fig. \ref{si_fig:controlEff_pickplace}). The results are shown in Fig. \ref{fig:switch_method}d, where multiple trials are plotted (grasping and placing a blue paper onto a yellow/purple plate). 
The time axes of each trajectory are aligned to the average execution duration of each control period for better comparison.
In the planar x-y space, both the VLA model exhibit stable control effort with low variance during arm motion. While the dexterous low-level controller increases control effort while adjusting the arm position prior to grasp, followed by a decrease during the dexterous grasping period.
For the control effort of finger joint $\dot x$, the low-level controller produces structured fluctuations with similar patterns across trials, indicating the execution of similar dexterous grasping motions. During the VLA-controlled phases, finger control effort remains as zero, as the VLA model does not command finger-level motions. These results quantitatively confirm the complementary roles of the high-level and low-level controllers and demonstrate effective coordination between them.

\subsubsection{Language-Conditioned Pick-and-Place Tasks}
To validate the language conditioned manipulation capabilities of the proposed switching-based framework, we evaluate its performance on language-conditioned pick-and-place tasks and compare it with a VLA-only baseline. The task involves language-conditioned pick-and-place of three representative objects with varying geometries, a red pepper, a roll of tape, and a thin piece of blue paper, each to be placed onto one of two target locations (a yellow or purple plate; Fig. \ref{fig:switch_method}c). Two objects are presented simultaneously: either pepper and tape, or pepper and blue paper.
The object and target plate are specified via natural language instructions provided to the VLA model.

The framework is evaluated across all object-placement combinations, with five trials conducted for each condition. Single VLA model is used for high-level arm motion and different low-level diffusion policies are trained for grasping different objects. For comparison, we also evaluate a baseline that relies solely on a VLA model trained to perform the full pick-and-place process. In this baseline, grasping is executed using the pre-programmed action sequence employed during VLA data collection, which encodes the robot hand's 13-DoF finger motion into a 1-DoF grasping proxy. To assess task performance beyond a binary success metric, we adopt a graded success score based on task-specific criteria, as detailed in Sec. \ref{sec-sub:scoringrules}.

Fig. \ref{fig:switch_method}e reports the success scores for each object-placement combination. For the VLA-only baseline, the model is trained exclusively on the pepper and tape objects, as the simplified 1-DoF grasping strategy is not capable of grasping the paper. Under this baseline, the VLA-only approach achieves average success scores of 0.43 for the pepper and 0.23 for the tape. In contrast, the proposed switching-based framework combining the VLA model with the low-level diffusion controller achieves substantially higher performance, with average success scores 0.82 across all object–placement combinations. Performance remains consistent across objects with different geometries, indicating stable grasp execution using object-specific dexterous hand configurations. Slightly lower scores are observed when placing objects onto the yellow plate compared to the purple plate, which can be attributed to the smaller target area and increased likelihood of placement outside the plate. Overall, these results demonstrate the advantages of integrating a high-level VLA model for language understanding and task planning with a low-level diffusion policy for precise and dexterous manipulation.

\subsubsection{Recovery Behavior \& Event Signal Robustness}
Recovery from failures is critical in complex manipulation tasks. In our switching-based framework, the event signal enables robust recovery behavior by preventing premature controller switching and indicating true subtask completion. Fig. \ref{fig:switch_method}f–\ref{fig:switch_method}g illustrate a recovery scenario from a failed grasping and the corresponding event signal $\sigma$ response.

In this example, the robot attempts to grasp a thin piece of paper but fails to achieve a stable grasp on the first two attempts. These attempts are reflected by three successive increases in the metacarpophalangeal (MCP) joint angle of the index finger, shown by the blue trace in Fig. \ref{fig:switch_method}g, corresponding to repeated finger closing motions. Notably, during the first failed grasp attempt, $\sigma$ exhibits a transient peak that exceeds the switching threshold but does not remain above the threshold for the required number of time steps to fill the switch condition. As a result, the controller switch is not triggered, allowing the low-level diffusion policy to automatically initiate a reattempt. During the second attempt, $\sigma$ remains near zero, again leading to another reattempt. Only on the third attempt, when a stable grasp is successfully achieved, does $\sigma$ rise above the threshold and remain consistently high, triggering the controller switch and allowing the task to progress. This behavior demonstrates both the recovery capability of the diffusion-based controller and the temporal stability of the event signal, which together enable robust execution without premature termination or switching.

\begin{figure*}[tb]
     \centering
     \includegraphics[width=0.98\textwidth]{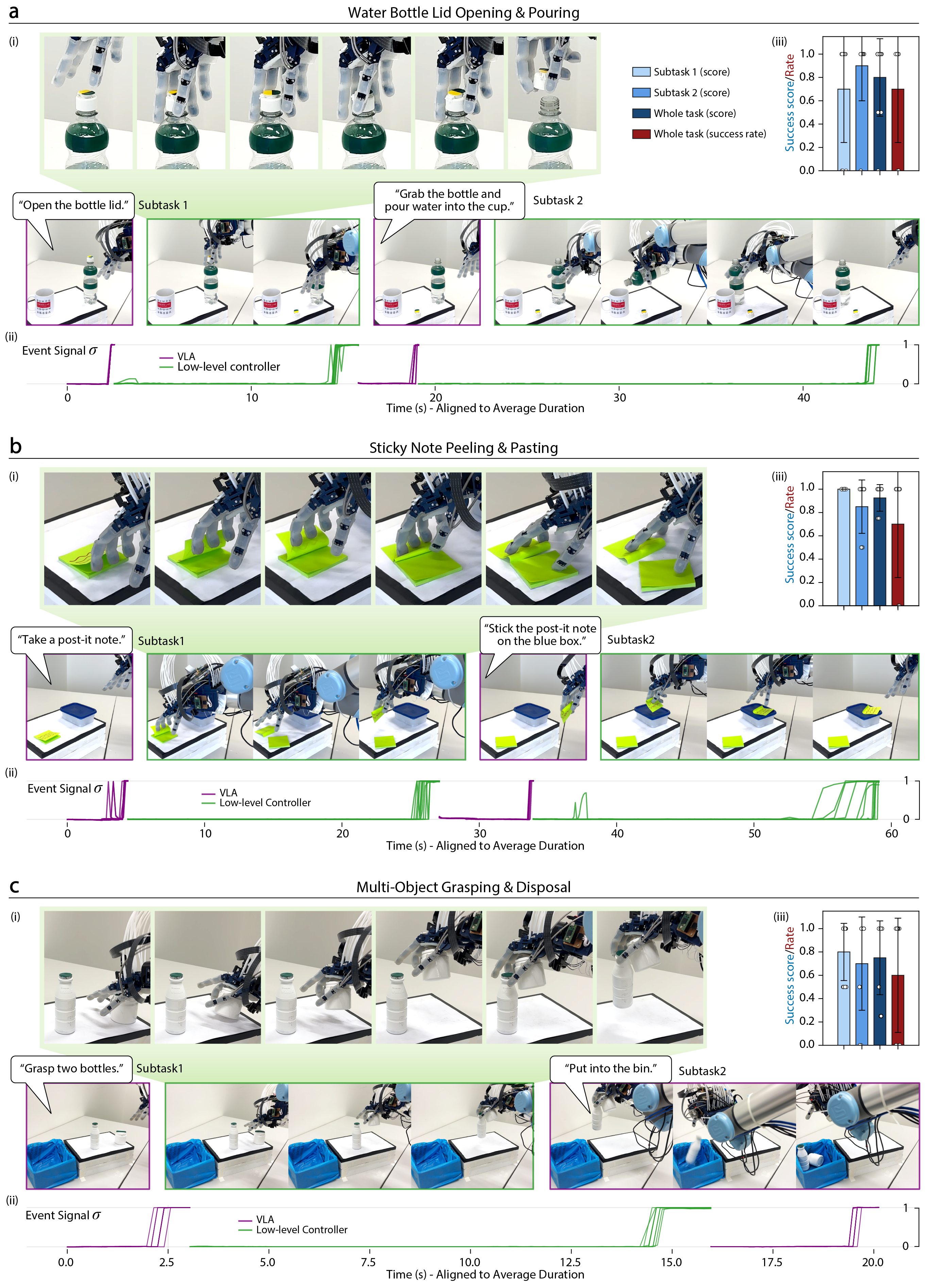}
     \caption{Snapshots of three dexterous manipulation tasks, reported success score and rate, and event signal responses. Single VLA model with different low-level controllers are used across these tasks (Supplementary Video 5).
     }
     \label{fig:dex_demos}
\end{figure*}

\subsection{Multi-Step Dexterous Task Demonstrations}
\label{sec:dexDemos}

To demonstrate the effectiveness of the proposed switching-based framework in highly dexterous manipulation, we evaluate it on three tasks: water bottle lid opening \& pouring (dex-task A), sticky note peeling \& pasting (dex-task B), and multi-object grasping \& disposal (dex-task C), shown in Fig. \ref{fig:dex_demos}.  We pick these tasks as they leverage our pipeline and dexterous compliant hand as they require contact-rich interaction with the environment.

Each task consists of two subtasks requiring both global arm motion and precise finger-level manipulation. In the dex-task A, the robot first unscrews a bottle cap and then pours water into a cup. The dex-task B involves separating a single post-it note from a stack and pasting it onto a box lid. The dex-task C requires picking multiple objects in a single grasp and placing them into a bin.
The dexterous components of these tasks are particularly challenging. For example, unscrewing a bottle cap requires coordinated force application from all fingers to generate sufficient frictional torque while stabilizing the bottle. Separating a post-it note involves applying controlled normal force and lateral sliding to peel a single sheet from the stack. For multi-object grasping, the robot must maintain a stable grasp on one object using part of the hand while simultaneously using other fingers to grasp an additional object.

In our framework, a single VLA model is used for all tasks and different low-level diffusion policies are trained for each subtask. Each subtask is executed through sequential control by the VLA model followed by the low-level controller. The VLA model first move the hand to the target object, after which the low-level controller performs the dexterous manipulation. An exception is the “Put into the bin” step of dex-task C, which only requires object transport and release. After the switching is activated in this step, it triggers a hand opening action to release the object instead of using a low-level controller considering the simplify of releasing motion (same as releasing step of language-conditioned pick-and-place task in Sec. \ref{sec:coordinate}).
As a result, the control sequence follows a VLA–lowlevel–VLA–lowlevel pattern for the dex-task A and B, and a VLA–lowlevel–VLA pattern for the dex-task C.

Fig. \ref{fig:dex_demos} presents the timeseries images of the task, corresponding performance evaluation, and event signal response illustrating the switching activities between the VLA model and the low-level controller.
The event signals for each trial are shown in Fig. \ref{fig:dex_demos}a(ii), b(ii), c(ii), with time axis aligned to average execution duration of each control period for better illustration. The results demonstrate that the predicted event signals reliably track subtask completion and enable correct switching between controllers across trials.
We quantitatively evaluate the performance using a task-specific success score (see Sec. \ref{sec-sub:scoringrules} for metrics). The mean and standard deviation over 10 trials are shown in Fig. \ref{fig:dex_demos}a(iii), b(iii), c(iii). For each task, the left three bars represent the success scores of individual subtasks and the overall task, while the final bar indicates the full-task success rate (i.e., achieving full score across all subtasks). All tasks achieve success rates above 0.6. Failure cases primarily arise from the difficulty of precise finger–object interaction. Common failure modes include unsuccessful cap opening in the dex-task A, dropping the post-it note during placement or placing it with the note side facing downward in the dex-task B, and losing the first object while attempting to grasp a second object in the dex-task C.

Although the presented tasks involve two subtasks and a limited number of controller switches, the proposed switching-based framework naturally extends to longer-horizon manipulation tasks requiring additional subtasks and more switching activities.

\begin{figure*}[tb]
     \centering
     \includegraphics[width=0.95\textwidth]{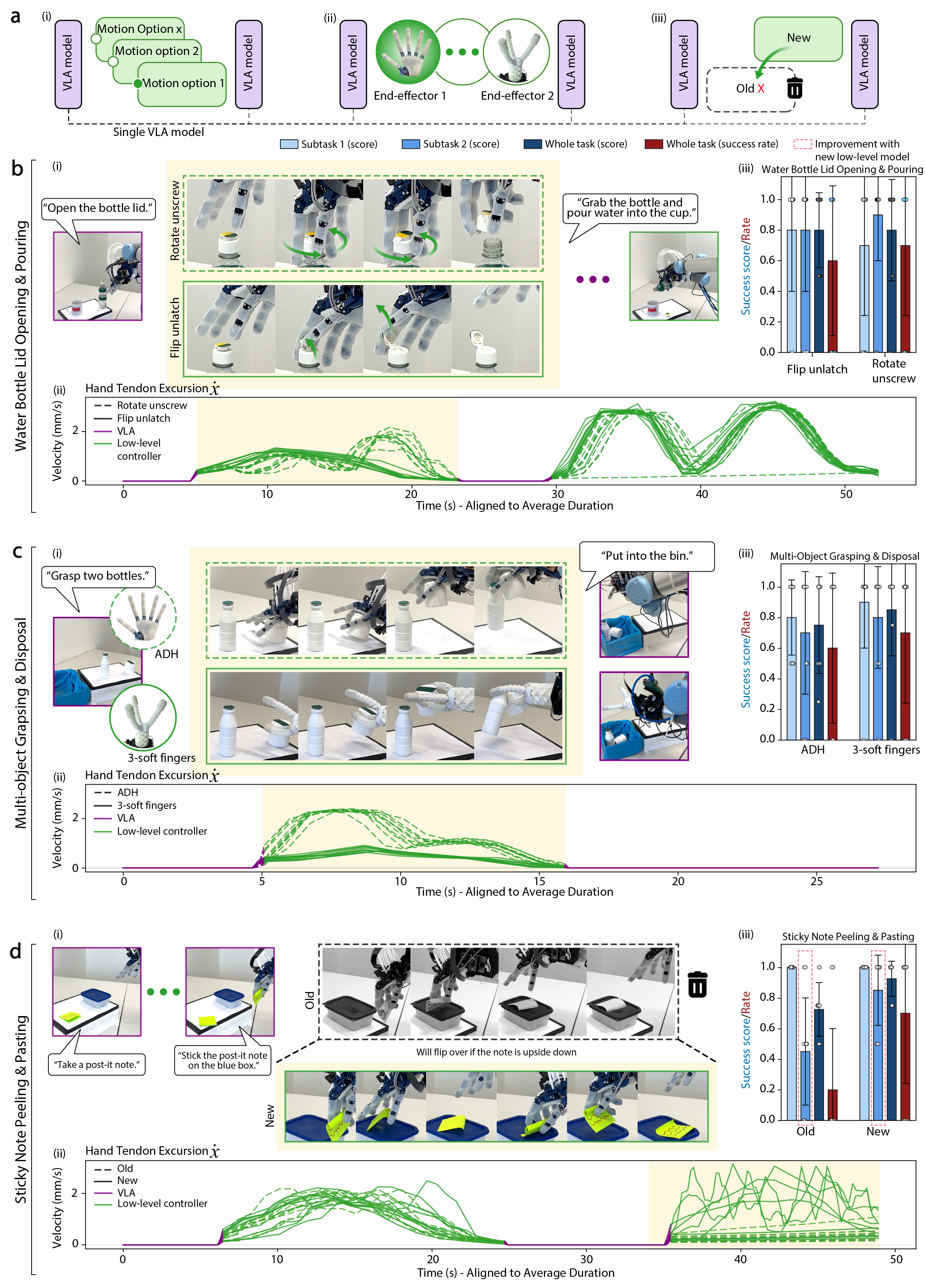}
     \caption{Model adaptation via modularity of switching-based dual-channel framework. a, Overview of different adaptation cases. b, Adaptation to different dexterous motion behavior. c, Adaptation to different dexterous manipulators. d, Improvement of subtask performance via component-wise update. For each adaptation case, we report a sequence of keyframes, control effort trajectories of the dexterous manipulator (more results about arm translation and orientation are shown in Supplementary Fig. \ref{si_fig:controlEff_rotateflip}-\ref{si_fig:controlEff_postit}), task success scores and success rate. (Supplementary Video 6)}
     \label{fig:modular}
\end{figure*}

\subsection{Model Adaptation via Modularity}

Our switching-based framework facilitates efficient model adaptation and generalization through its inherent modularity. The modularity ensures that the modification or update of the dexterous low-level controller does not affect the high-level VLA model, eliminating the need to retrain or adapt the VLA and thereby significantly reducing training data requirements and computational cost. There are three specific uses and applications of modularity (Fig. \ref{fig:modular}a). The first is to enable different dexterous motion behaviors, whilst keeping the high level, visual planning the same. Secondly, it allows rapid adaptation to significantly different dexterous hardware. Finally, where there is a particular sub-task which is the lowest performing, we can retrain just that component (low-level policy), without risking degradation in the performance of other subtasks.
We now present three representative studies illustrating these three capabilities (Fig. \ref{fig:modular}). 


\paragraph{Case 1: Behavior Change in Water Bottle Lid Opening.}
In the dex-task A, we modify the lid-opening behavior from rotational unscrewing and removing the cap to directly pushing up and unlatching a flip-top lid (see Fig. \ref{fig:modular}b(i)). Instead of retraining the full system, we train a new low-level controller specialized for the unlatching behavior and replace the original controller for this subtask. This demonstrates that behavior modalities can be flexibly substituted based on task requirements or user preferences without retraining other components and the risk of degrading performance in other subtasks, especially the computationally expensive VLA model.
The control effort profile (Fig. \ref{fig:modular}b(ii)) reflects this behavioral shift: the “flip unlatch” motion exhibits less fluctuation compared to the more complex “rotate unscrew” motion, due to simpler finger coordination. Notably, the control effort for the subsequent pouring subtask remains almost unchanged. The success score and success rate (Fig. \ref{fig:modular}b(iii)) are comparable across both behavior modalities, demonstrating that the modification preserves overall task performance.

\paragraph{Case 2: Adaption to Different Dexterous Hardwares}

To show the adaption to different hardwares, we show the dex-task C across two hardware platforms.  Our ADAPT hand and also a compliant, 3 soft fingered dexterous gripper \cite{guan2025dexterous} (Supplementary Fig. \ref{si_fig:softfingers}) which has 6 DoF.  This new gripper has different kinematics, degrees of freedom, visual appearance and grasping behaviors.  A new low-level controller is trained similarly through data collected from teleoperation, but only for this grasping step. The finger and wrist motion of the human which is now mapped to the action space of the 3 fingered grippers.  For the new 3 fingered gripper,  the same VLA is used for approaching and objects transportation motion, although the VLA is never trained on the data of 3 fingered gripper.
With this we observe that the success score and rate of the 3 fingered gripper is around 10\% higher than those of ADH, showing the advantages of this gripper design for multi-object grasping tasks. The success score of Subtask2 ("put into the bin", executed by old VLA model) is also comparable to that of ADH, showing the generalization ability of VLA to different hardwares.

\paragraph{Case 3: Performance Improvement in Sticky Note Pasting.}
In the dex-task B, the original policy exhibits unsatisfactory performance in the second subtask (“pasting”), primarily due to failure cases where the note is placed upside down, such that the side containing the notes faces downward. Because such correction behaviors are absent from the demonstration dataset, the original controller lacks the capability to detect and recover from this error. To address this, we augment the training dataset with demonstrations that include corrective flipping actions (flipping the note when placed incorrectly) and retrain only the corresponding low-level controller. As shown in Fig. \ref{fig:modular}d(i), the new controller successfully incorporates recovery behaviors. The control effort for subtask 2 exhibits increased variability, reflecting the more complex finger motions required for flipping, sometimes involving multiple attempts. In contrast, the control effort for the subtask 1 remains unaffected. Quantitatively, the success score for the second subtask improves from 0.45 to 0.85, yielding a 0.2 increase in whole task success score and a 0.5 improvement in whole task success rate.

These results highlight that the modular switching-based framework enables localized improvements and behavioral adaptation at the subtask level, while preserving the performance of unaffected components, and further highlights the generalization capability of the VLA. This eliminates the need for end-to-end retraining, offering a scalable and computationally efficient pathway for continual system adaptation and refinement.

\section{Discussion}\label{sec3}

We present a switching-based control framework that enables the deployment of VLA models on high-DoF dexterous manipulators. By integrating a pretrained VLA models with low-level diffusion-based controllers, the system supports language-conditioned dexterous manipulation across a range of complex tasks. Training data is collected via teleoperation using a compliant robotic hand, allowing demonstrations that capture safe contact rich interactions. We show that hardware-level compliance plays a critical role in improving manipulation robustness. Quantitative evaluations demonstrate that compliance enhances tolerance to both internal control disturbances and external variations in object properties and environmental conditions. Furthermore, the low-level controller is capable of learning multimodal manipulation strategies from demonstrations. However, the performance is sensitive to the initial configuration of the hand, which must be appropriately initialized by the VLA model. This interdependence highlights the necessity of combining high-level planning with low-level dexterous control, and enabled us to analyze the complementary roles of the VLA and low-level controllers through their control effort profiles, revealing distinct responsibilities in arm-level transport and finger-level dexterous manipulation. A learned event signal is introduced to monitor subtask progression and to govern switching between controllers. We demonstrate that this signal is robust to grasping failures and recovery attempts, and reliably predicts true subtask completion, and validate it across a range of dexterous tasks. We further demonstrate the capability the switching-based framework on diverse dexterous tasks, and showing the advantage of its modularity for efficient model updates and adaptation across different dexterous behaviors and hardware embodiments.




There remains a number of open and future questions. While a single VLA model generalizes across tasks, separate low-level controllers are currently required for different subtasks. Future work could explore sharing controllers across subtasks with similar motion primitives to improve data efficiency and generalization, although this may reduce the modular independence of the framework and requires careful trade-off analysis.  Moreover, recovery behaviors are presently confined within individual subtasks. If the event signal prematurely triggers a transition, the system cannot revert to a previous subtask to correct the failure, potentially leading to cascading errors in following subtasks. This could be addressed by extending the capabilities of the VLA so that it can verify subtask preconditions and dynamically re-plan, enabling cross-subtask recovery and more robust long-horizon task execution.  Another way to increase robustness in through integration of sensing. In the tasks we explore we see largely that the camera images are sufficient to capture physical interaction and contact without relying on tactile sensors.  However, for many tasks, such as those that are with transparent material, or more complex or compliant interactions will require tactile and more precise proprioceptive sensing. In our framework, we could integrate sensing into only the low-level control policies to limit overall complexity and still leverage the advantages of pre-trained large AI models. 
There is also a fundamental question regarding this approach of leveraging extensive demonstrations and data through teleoperation for developing controllers, both for VLA and the dexterous low-level controller. Whilst this is a practical setup, this has limits in generalization and adaption to out of domain environments and tasks.  Alternative approaches such as reinforcement learning come with their own challenges including sim2real transfer and reward tuning, however, they could offer a means of removing the need for human-data capture. An intermediate step of imitation learning, and bootstrapping these policies with reward learning and reinforcement learning could be a practical way to relative advantages, but moving slowly to a more generalizable solutions.

\section{Methods}\label{sec4}

\subsection{ADH Robot Hand with Variable Compliance}




We introduce the ADH robotic hand, focusing on its spring-extension sensing, compliant and rigid operation modes, onboard vision, and communication architecture (Supplementary Fig. \ref{si_fig:ADH}). In this work, we extend the robotic hand introduced in \cite{junge2025adapt} (ADAPT Hand 2, ADH) by incorporating spring-sensing capabilities while preserving its human-matched kinematics and passive dynamics. Similar to its predecessor, each metacarpophalangeal (MCP) joint of the four fingers and the two carpometacarpal (CMC) joints of the thumb are implemented as series elastic joints. The compliance arises from springs located at a centralized carriage mechanism. Each spring carriage is equipped with an independent potentiometer to measure spring displacement (Fig. \ref{fig:hardwares}b), enabling direct sensing of elastic deformation. Potentiometer readings are acquired via a Teensy 4.0 microcontroller, which communicates serially with a Raspberry Pi 4B mounted on the hand.

When a finger establishes contact with the environment, external forces induce spring extension; thus, spring displacement serves as a proxy for contact estimation. However, estimating contact purely from spring extension presents challenges. Friction in the actuation system, arising from tendon routing, bowden cables, and bearings, can cause spring deflection even in the absence of environmental contact. Nevertheless, as illustrated in Fig. \ref{fig:hardwares}d,  contact interactions typically result in larger and more sustained spring extensions, allowing partial observation for finger-object contact.

The introduction of series elasticity enables operation in a compliant configuration, improving tolerance to contact uncertainties. For applications requiring higher structural stiffness, compliance can be mechanically disabled. The rigid configuration is achieved by inserting a locking component into the carriage assembly, physically blocking spring deformation.

Hand proprioception is obtained from motor encoder readings, which are transmitted to the same Raspberry Pi 4B used for spring sensing. In addition to proprioceptive measurements, the hand is equipped with three onboard cameras to capture detailed finger–object interactions (Supplementary Fig. \ref{si_fig:ADH}): Cam 1 (Dorsal Camera): Mounted on the back of the hand and angled forward to monitor the dorsal side of the fingers and fingertips. Cam 2 (Lateral Palmar Camera): Positioned laterally on an extended beam, providing an oblique view of the palm and finger surfaces. Cam 3 (Central Palmar Camera): Mounted at the base of the palm, oriented upward toward the inner palm and the undersides of the fingers. All visual streams are processed by an additional Raspberry Pi 4B mounted onboard the hand, enabling synchronized acquisition of multimodal sensory data. Note that in the language-conditioned pick-and-place experiments, an earlier version of the ADH was used. This version includes only a single wrist-mounted camera, located in the same position as Cam 3 (Central Palmar Camera). All other components remain unchanged.

\subsection{Shared Control for Teleoperation}


We explain the shared control scheme to enable more easy teleoperation and high quality data. 
For teleoperation, a VR device is used to track the teleoperator’s hand keypoint (finger tips and joints) positions. This representation is then converted into finger joint angles, and mapped to motor commands that control tendon excursions, thereby controlling joint movements of robot hand. 
The shared control scheme uses a state machine to switch between free teleoperation and autonomous force control based on contact detection via spring displacement/extension. For the four fingers, when the spring displacement exceeds a threshold, the finger enters the force control state where the controller maintains a target spring displacement (force), where the operator’s joint angle at contact is recorded, and the system remains in force control until the operator extends the finger enough to reduce displacement below the threshold, returning to the off state. This prevents the finger from staying permanently in control and allows seamless transitions between free motion and assisted grasping. The thumb follows a different logic: it only enters the control state when it is close to another fingertip that is already in control, reflecting its opposable role in grasping. In control mode, the thumb’s motion direction is determined by the relative magnitudes of its two force setpoints (target value that a controller tries to maintain), which are computed from the relative distances between the thumb tip and the fingers. For every finger and thumb, when a transitions into control, its other joints are locked, ensuring stable and coordinated grasp assistance.

\subsection{Compliant 3 Fingered Gripper}

We introduce the 3-soft finger dexterous gripper \cite{guan2025dexterous} (Supplementary Fig. \ref{si_fig:softfingers}) used to demonstrate the generalization ability of our framework across different embodiments. This gripper has 6 DoF in action space, and triggered by corresponding 6 tendons and motors, two for each finger, controlling its bending direction and magnitude. Its compliant trimmed helicoids (TH) structure and the softness of the materials (thermoplastic polyurethane, TPU) enable the passive compliance characteristics, improving the robustness of object-finger contact during dexterous manipulation.
For teleoperation, the human finger tip motion of thumb, index finger and pinky finger is capture by the VR device and mapped to 3-soft robot fingers. Each soft robot finger has 2 DoF and can bend towards other the two robot fingers. After an initial calibration that records the initial positions of the three human fingertips. Two unit direction vectors are defined for each finger, pointing to the other fingertips. The displacement of each human fingertip is then computed relative to its initial position and projected onto these two directions. The resulting projections determine the bending amplitudes of each finger, corresponding to its 2 DoF bending motions. 
To capture visual information from a wrist perspective, a wrist camera is amounted on the 3-finger gripper to capture the finger motions. During data collection, image from a third-person camera, the wrist camera, and all motor positions (proprioception) are recorded. This compliant 3-finger gripper is used in dex-task C to demonstrate the generalization and adaptation ability of our switching-based framework.

\subsection{Experimental Setups}
As shown in Fig. \ref{fig:hardwares}a, the ADH hand is mounted on a UR5 robotic arm. A third-person camera positioned on a tripod captures a global view of the arm, hand, and workspace objects. The wrist-mounted cameras on the ADH provide close-up visual observations of finger motions and finger–object interactions.

The objects used across all experiments are summarized in Supplementary Fig. \ref{si_fig:exp_obj}. In the pick-and-place tasks, we use an artificial red pepper, a roll of tape, a blue sheet of paper, and a thin square block. 
For the knob-turning experiments, we use a custom knob apparatus with three interchangeable knob sizes (45 mm, 55 mm, and 65 mm in diameter). The rotation angle of the knob is measured using an AS5048B magnetic encoder and processed by an Arduino Uno, which transmits the measurements externally via a USB connection.
In the dex-task A, a water bottle and a cup are used. To improve visual observability, the transparent bottle is wrapped with green tape, and the cap is marked with a split yellow–green label to make rotational motion visually distinguishable. The bottle is sealed to prevent water from actually being poured during the experiments.
For the dex-task B, we use a stack of bright green sticky post-it notes and a blue box. The bottom edge of the top note is intentionally curled upward to create a small gap, allowing the robotic hand to more easily slide  underneath and peel off a single sheet.
In the dex-task C, two white bottles and a blue trash bin are used. Colored markers are attached to the bottle caps to improve visual contrast against the white background.

\subsection{Demonstration Data Capture}
Demonstration data including robot arm motion, dexterous hand motion, and event signals for imitation learning are collected via teleoperation using an VR device and a gamepad interface. Visual data (images) is recorded by the thrid-person camera and wrist cameras. During teleoperation, the teleoperator explicitly segments demonstrations into arm transport and dexterous manipulation phases by pressing a button on the gamepad. These annotations are used to separate trajectories: arm motion segments are used for VLA fine-tuning, while dexterous hand motion segments are used to train the low-level controller. To ensure smooth transitions between controllers, we include a temporal overlap of five steps before and after each segmented trajectory. This overlap can improve continuity and robustness during switching.

For event signal learning, the final 10 timesteps of each segmented trajectory are labeled as 1, with all preceding steps labeled as 0. This labeling strategy encourages the model to predict a rising event signal as the subtask approaches completion.

For language-following pick-and-place tasks, 20 repeated demonstration trajectories are recorded for every object and placement combination. For dex-task A-C, we collect 30 demonstration trajectories for per task. The sampling frequency is around 5 Hz. Despite the relatively small dataset size, the proposed framework achieves strong performance, highlighting its data efficiency.


\subsection{Model Training and Implementation}
Our approach is data-efficient: instead of retraining VLA models for dexterous manipulation, we fine-tune them with a small amount of data to predict event-based switching signals, while learning lightweight dexterous controllers at the subtask level.
Segmented arm motion trajectories (6-DoF) are used to fine-tune pretrained VLA models. We adopt OpenVLA \cite{kim2024openvla} for pick-and-place tasks and $\pi 0$ \cite{black2024pi_0} for more complex dex-task A-C. We train a single VLA model across dex-task A-C. The original outputs of both models include end-effector pose (position and orientation) and a 1-DoF gripper command. In our framework, the gripper command is replaced with the learned event signal, using the annotated labels as supervision.
The VLA models take as input multi-view RGB observations (a third-person camera and a wrist camera (cam 1)), along with task-level language instructions. 

For low-level control, we train diffusion policies \cite{chi2025diffusion} using segmented dexterous manipulation trajectories, consisting of 15-DoF hand joint motions and corresponding 6-DoF arm motions. To improve robustness to visual variability, we apply standard data augmentation techniques, including random cropping and color and brightness perturbations. A separate low-level diffusion controller is trained for each subtask. At inference time, the appropriate controller is selected via a lookup table conditioned on the language instruction provided to the VLA model.
For simple release actions (e.g., object placement in pick-and-place task and object disposal in the dex-task C), we bypass policy execution and instead apply predefined open-loop commands to the manipulator for the releasing motion, given the low complexity of these motions.

\subsection{Event Signal-Driven Model Switching}

We enable the switching between the VLA model and the low-level controller based on the event signal $\sigma$. Control is initially handled by the VLA model. Given a language command, the VLA moves the robot hand toward the target object while simultaneously predicting $\sigma$ indicating the completion progress of the current subtask. When the predicted $\sigma$ satisfies the switching condition, where it exceeds a predefined threshold (0.8 in this work) for more than two consecutive inference steps, the switch mechanism is activated. This requirement helps prevent premature switching caused by noise, disturbances, or transient prediction errors. Once the switching condition is met, inference of the VLA model is stopped, and the corresponding low-level controller is activated. The low-level controller is selected according to the language command, which determines the current subtask. The low-level controller then performs the dexterous manipulation by controlling both the robot arm and the dexterous manipulator.

The same switching logic applies to the low-level controller. When its predicted event signal satisfies the same switching condition, the controller inference is terminated and control is returned to the VLA model. The VLA then proceeds with the next stage of the task based on the updated language command. This alternating control process continues until the final subtask is completed, as indicated by the event signal generated by the last low-level controller.





\bibliography{sn-bibliography}

\clearpage


\setcounter{section}{0}
\renewcommand{\thesection}{S\arabic{section}}

\setcounter{figure}{0}
\renewcommand\figurename{Supplementary Fig.}
\renewcommand{\thefigure}{\arabic{figure}}

\section*{Supplementary Information}

\section{Experiments Details and Scoring Rate}
\label{sec-sub:scoringrules}
To evaluate task performance beyond a binary success metric, we define task-specific scoring rules that capture partial completion of tasks.

\paragraph{Pick-and-place tasks}
The language instruction follows the format: ``Put (red pepper / tape / blue paper) on (yellow / purple) plate.'' Each object–plate combination is tested five times. Task performance is evaluated using a five-level score reflecting the degree of task completion:

\begin{itemize}
    \item 1.00 – Full task success (correct object grasped and placed on the correct plate).
    \item 0.75 – Correct object grasped but placed on the wrong plate.
    \item 0.50 – Correct object grasped but failed to place on a plate.
    \item 0.25 – Failed to grasp the correct object.
    \item 0.00 – The robot approaches or interacts with the wrong object.
\end{itemize}

\paragraph{More dexterous tasks}

\textbf{Water bottle lid opening \& pouring}
This task consists of two subtasks.

{Subtask 1:} Language instruction: ``Open the bottle lid.''  
The robot must open or unscrew the bottle cap.
\begin{itemize}
    \item 1 – Lid successfully opened.
    \item 0 – Lid not opened.
\end{itemize}

{Subtask 2:} Language instruction: ``Grab the bottle and pour water into the cup.''  
Because the bottle is sealed for experimental convenience, successful pouring is inferred indirectly. The attempt is considered successful if the robot grasps the bottle, tilts it such that the bottle spout is positioned above the cup area, and reaches a sufficient tilt angle that would cause water to accumulate near the spout.
\begin{itemize}
    \item 1 – Successful grasp and valid pouring posture.
    \item 0 – Otherwise.
\end{itemize}

\noindent \textbf{Sticky note peeling \& pasting}
This task also consists of two subtasks.

{Subtask 1:} Language instruction: ``Take a Post-it note.''
\begin{itemize}
    \item 1 – A single sticky note is successfully peeled from the stack and successfully grasped.
    \item 0 – Otherwise.
\end{itemize}

{Subtask 2:} Language instruction: ``Stick the Post-it note on the blue box.''
\begin{itemize}
    \item 0 – Note not placed on the box.
    \item 0.5 – Note placed on the box but with the side containing the adhesive facing downward (incorrect orientation).
    \item 1 – Note placed on the box with the adhesive side facing upward (correct orientation).
\end{itemize}

\noindent \textbf{Multi-object grasping \& disposal}
This task also includes two subtasks.

{Subtask 1:} Language instruction: ``Grasp two bottles.''
\begin{itemize}
    \item 0 – No bottle grasped.
    \item 0.5 – One bottle grasped.
    \item 1 – Both bottles grasped in a single grab.
\end{itemize}

{Subtask 2:} Language instruction: ``Put into the bin.''
\begin{itemize}
    \item 0 – No bottle placed into the bin.
    \item 0.5 – One bottle placed into the bin.
    \item 1 – Both bottles successfully placed into the bin.
\end{itemize}

For the three dexterous manipulation tasks described above, each task is evaluated over 10 trials. The overall task score of each task is computed as the average score across all subtasks. The task success rate is defined as the proportion of trials in which all subtasks achieve the full score.








\section{Extended Data and Results}\label{secA1}


\begin{figure*}[h]
     \centering
     \includegraphics[width=0.5\textwidth]{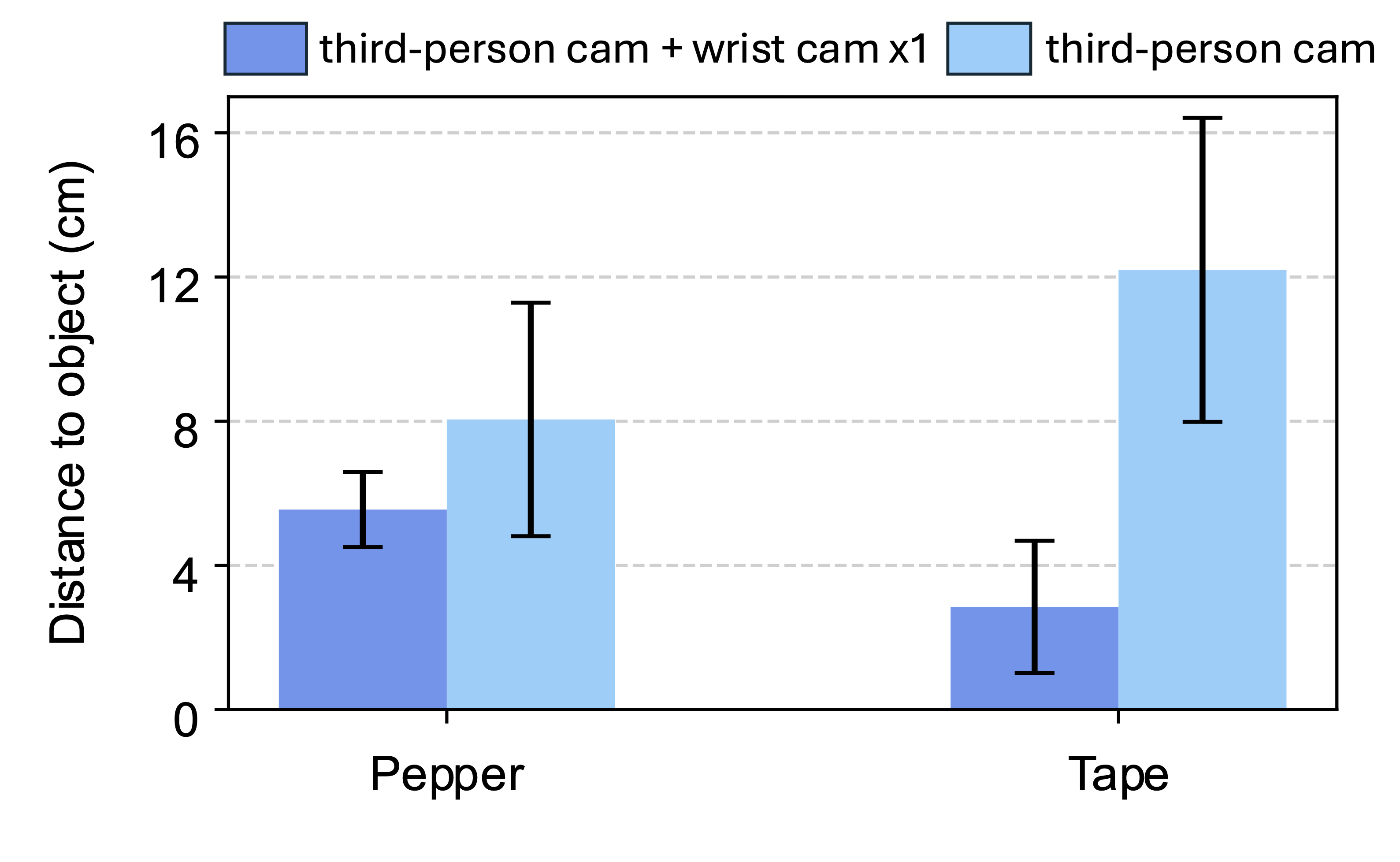}
     \caption{The x-y offset from the target object in the languge-conditioned pick-and-place task, when the VLA is used to approach an object. This is to evaluate the ’reaching’ precision of the VLA, the offset between the centre of the hand and the object is recorded in the x-y plane (i.e. parallel to the desk) after running the VLA. The position is recorded when the event signal is given by the VLA to switch to the low-level controller. 
     Results include when the VLA is trained with concatenated images of both third-person and one wrist camera view (cam 1) vs. VLA trained with single third-person camera view. Mean and standard deviation of five trials.}
     \label{si_fig:vla_dist}
\end{figure*}

\begin{figure*}[h]
     \centering
     \includegraphics[width=1.0\textwidth]{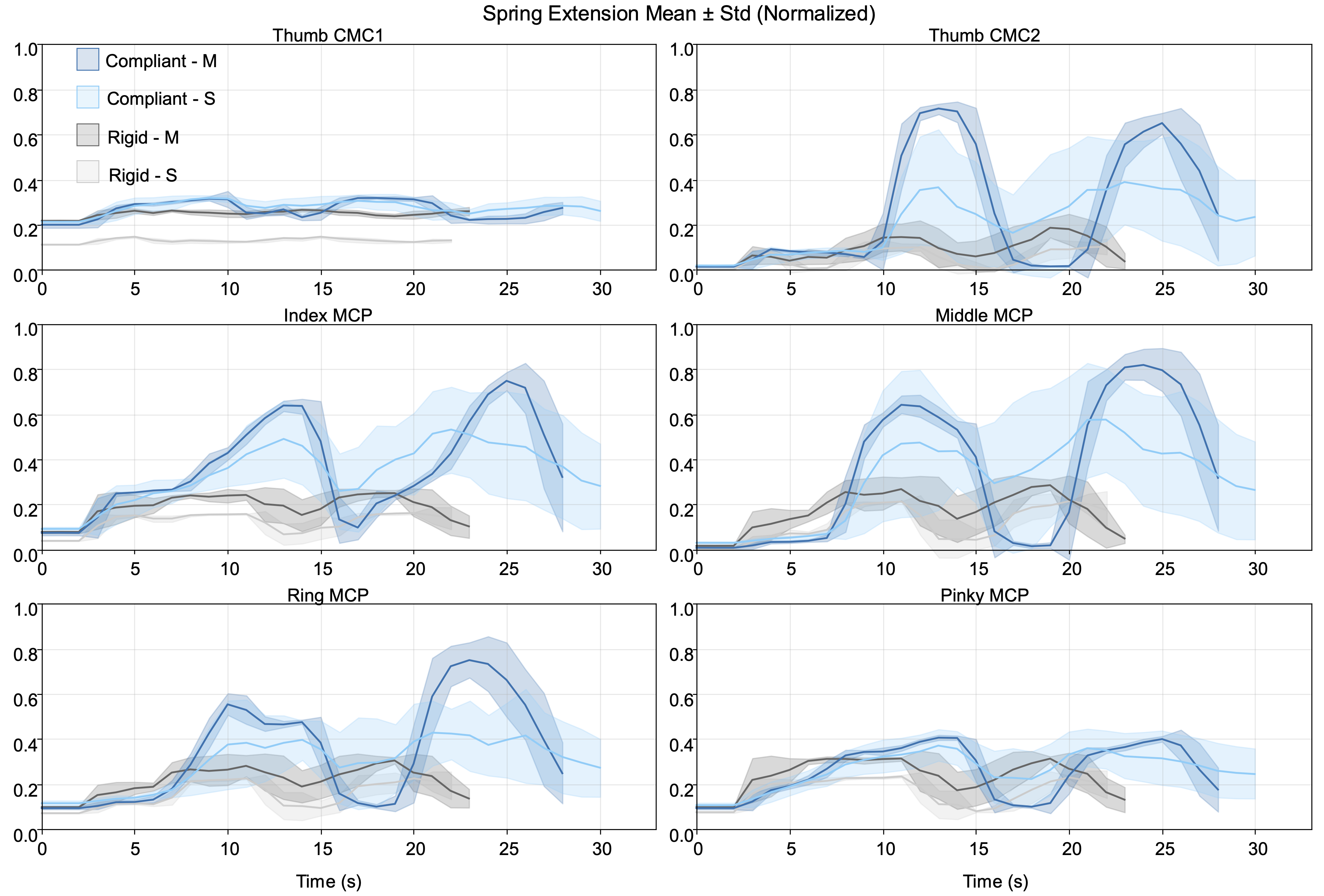}
     \caption{Measured spring extension of finger joints (under external disturbances - varying knob size), indicating adaptive behavior compliant robot hand. Supplementary results for Fig. \ref{fig:low-level_results}e.}
     \label{si_fig:spring_size}
\end{figure*}

\begin{figure*}[h]
     \centering
     \includegraphics[width=1.0\textwidth]{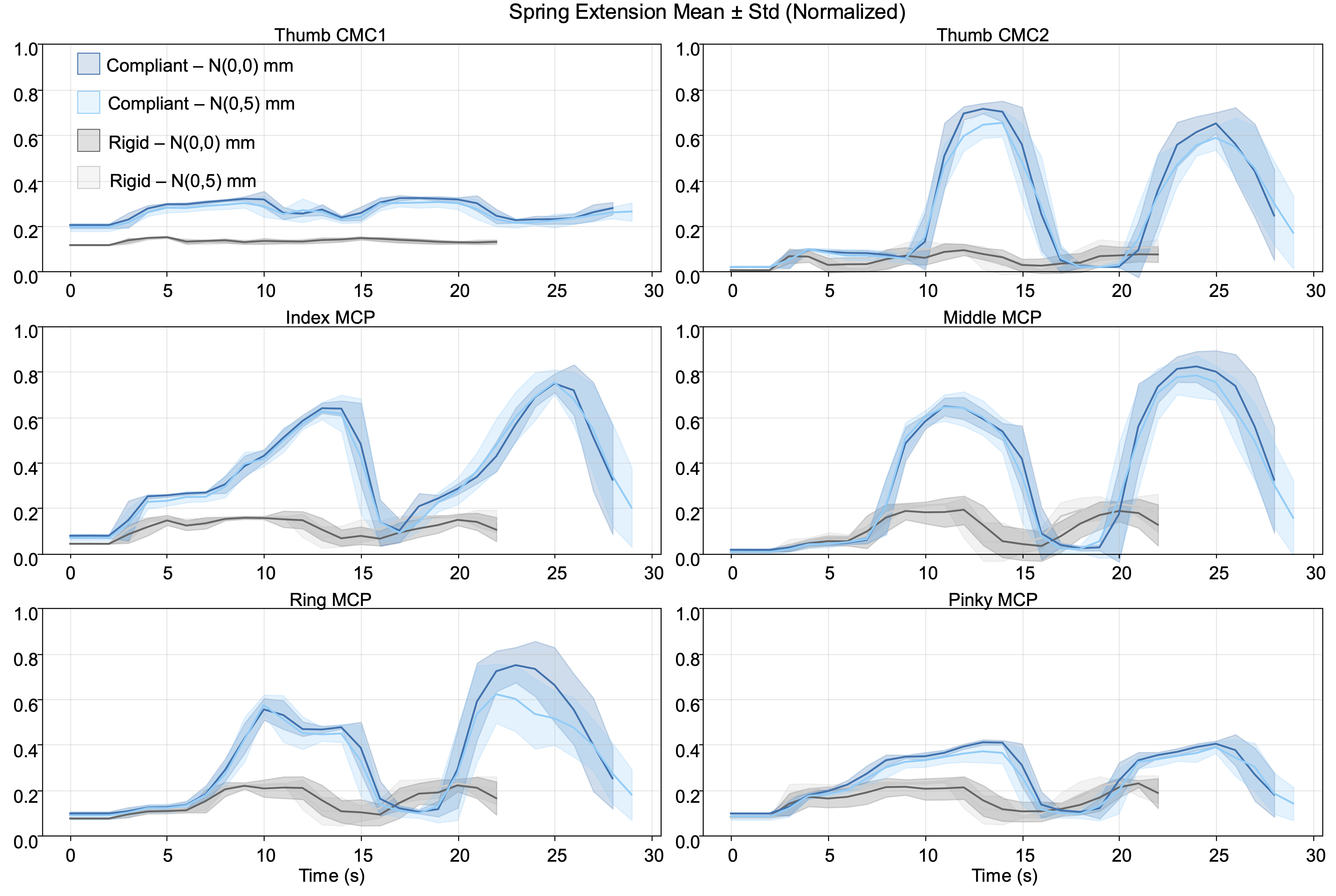}
     \caption{Measured spring extension of finger joints (under internal disturbances - simulated Gaussian control noise), indicating adaptive behavior compliant robot hand. Supplementary results for Fig. \ref{fig:low-level_results}e.}
     \label{si_fig:spring_noise}
\end{figure*}

\begin{figure*}[h]
     \centering
     \includegraphics[width=1.0\textwidth]{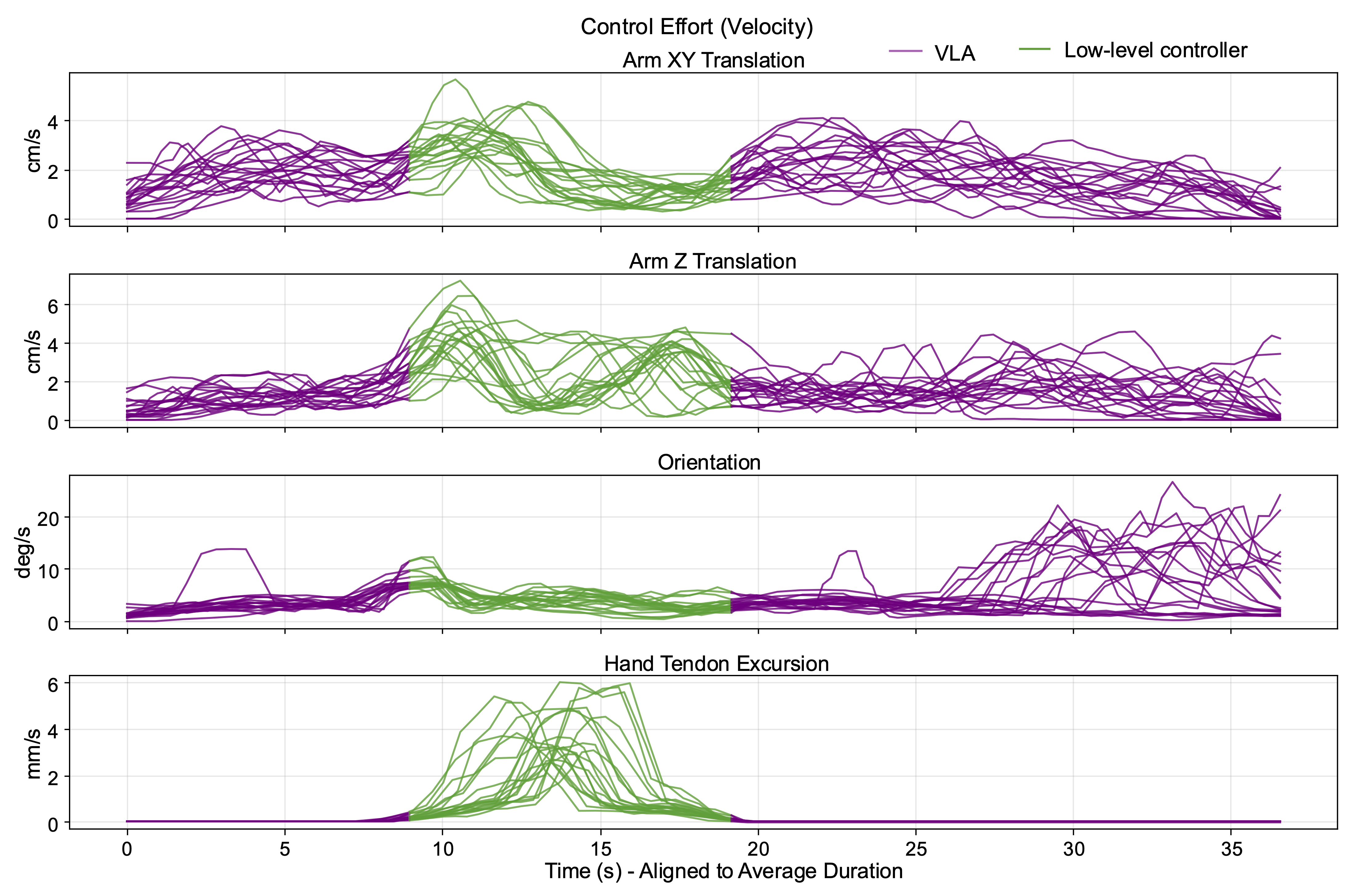}
     \caption{Control effort of coordinated models. Trajectories on language-conditioned pick-and-place tasks. Supplementary results for Fig. \ref{fig:switch_method}d}
     \label{si_fig:controlEff_pickplace}
\end{figure*}

\begin{figure*}[h]
     \centering
     \includegraphics[width=1.0\textwidth]{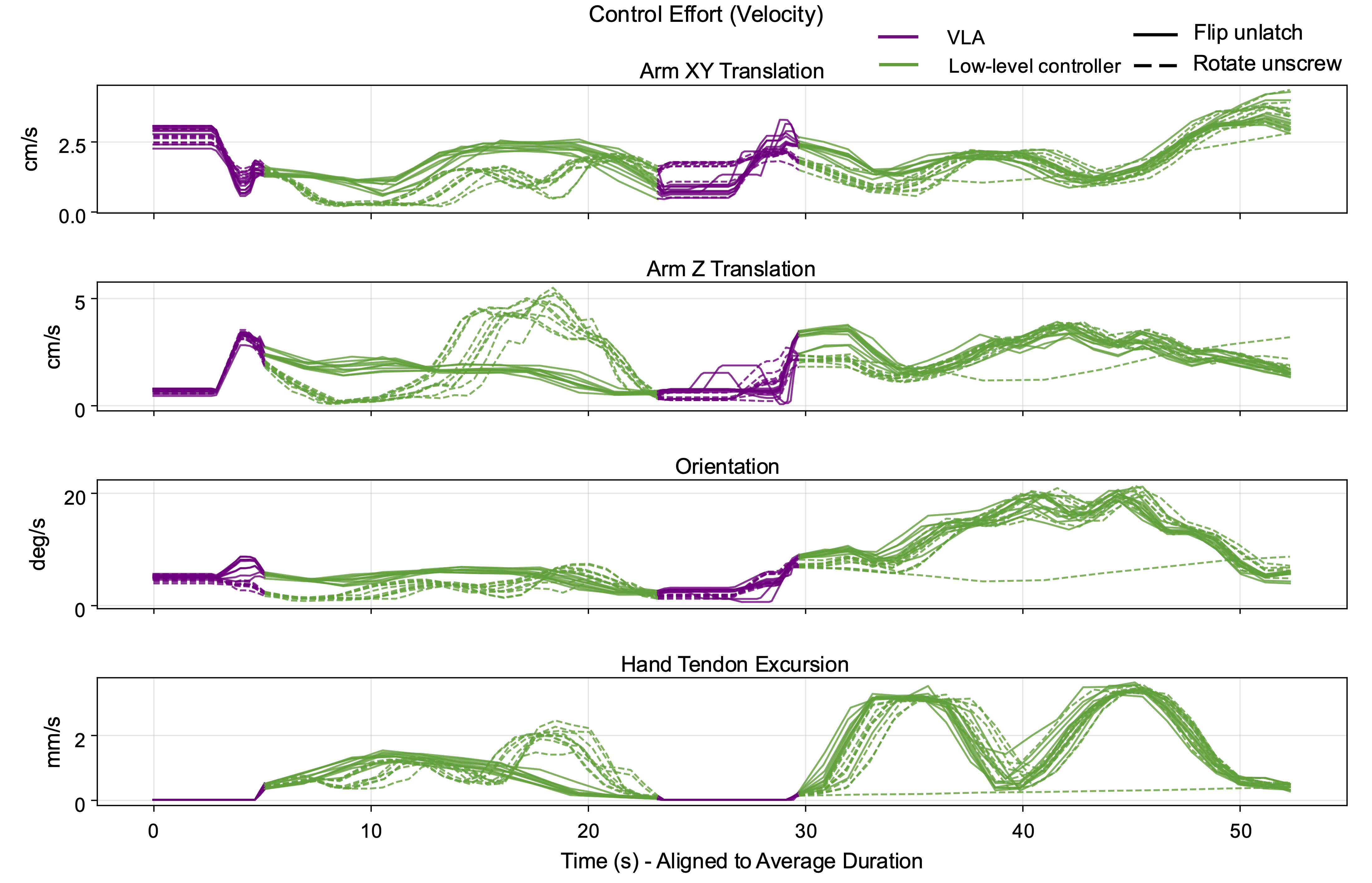}
     \caption{Control effort of coordinated models. Trajectories on water bottle lid opening \& pouring task (dex-task A). Supplementary results for Fig. \ref{fig:modular}b }
     \label{si_fig:controlEff_rotateflip}
\end{figure*}

\begin{figure*}[h]
     \centering
     \includegraphics[width=1.0\textwidth]{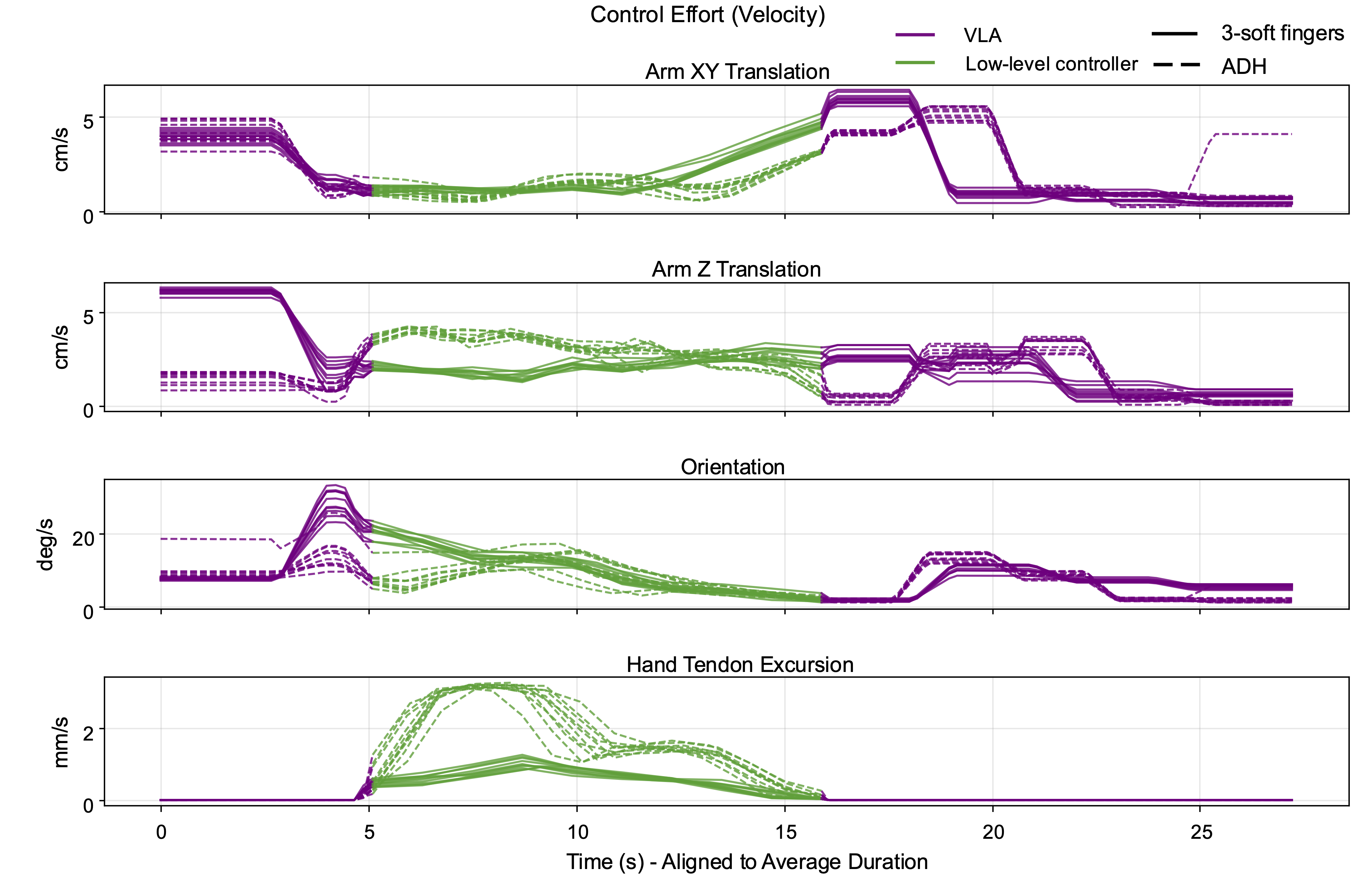}
     \caption{Control effort of coordinated models. Trajectories on multi-object grasping \& disposal task (dex-task C). Supplementary results for Fig. \ref{fig:modular}c }
     \label{si_fig:controlEff_softfingers}
\end{figure*}

\begin{figure*}[h]
     \centering
     \includegraphics[width=1.0\textwidth]{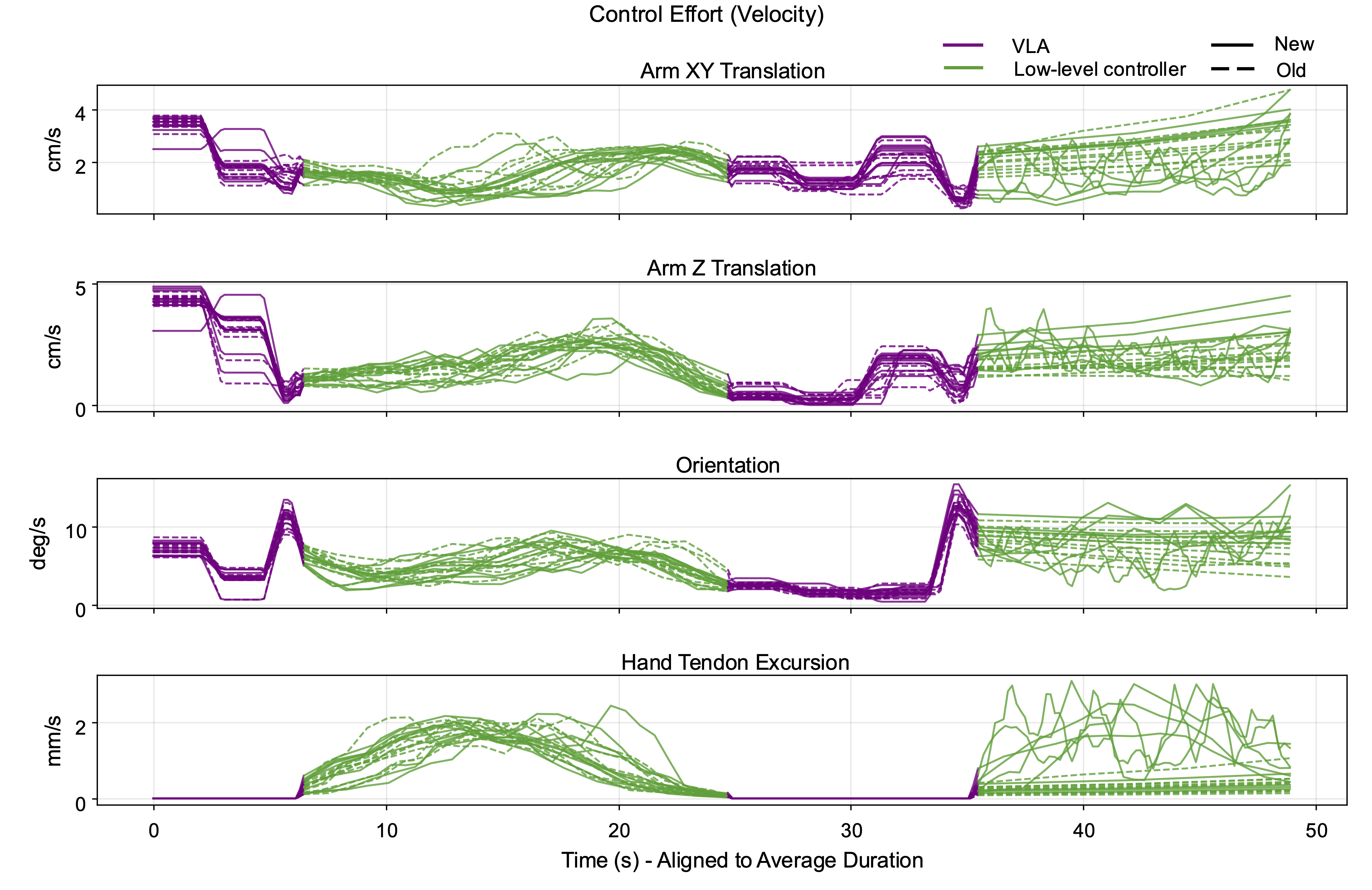}
     \caption{Control effort of coordinated models. Trajectories on sticky note peeling \& pasting task (dex-task B). Supplementary results for Fig. \ref{fig:modular}d}
     \label{si_fig:controlEff_postit}
\end{figure*}

\begin{figure*}[h]
     \centering
     \includegraphics[width=1.0\textwidth]{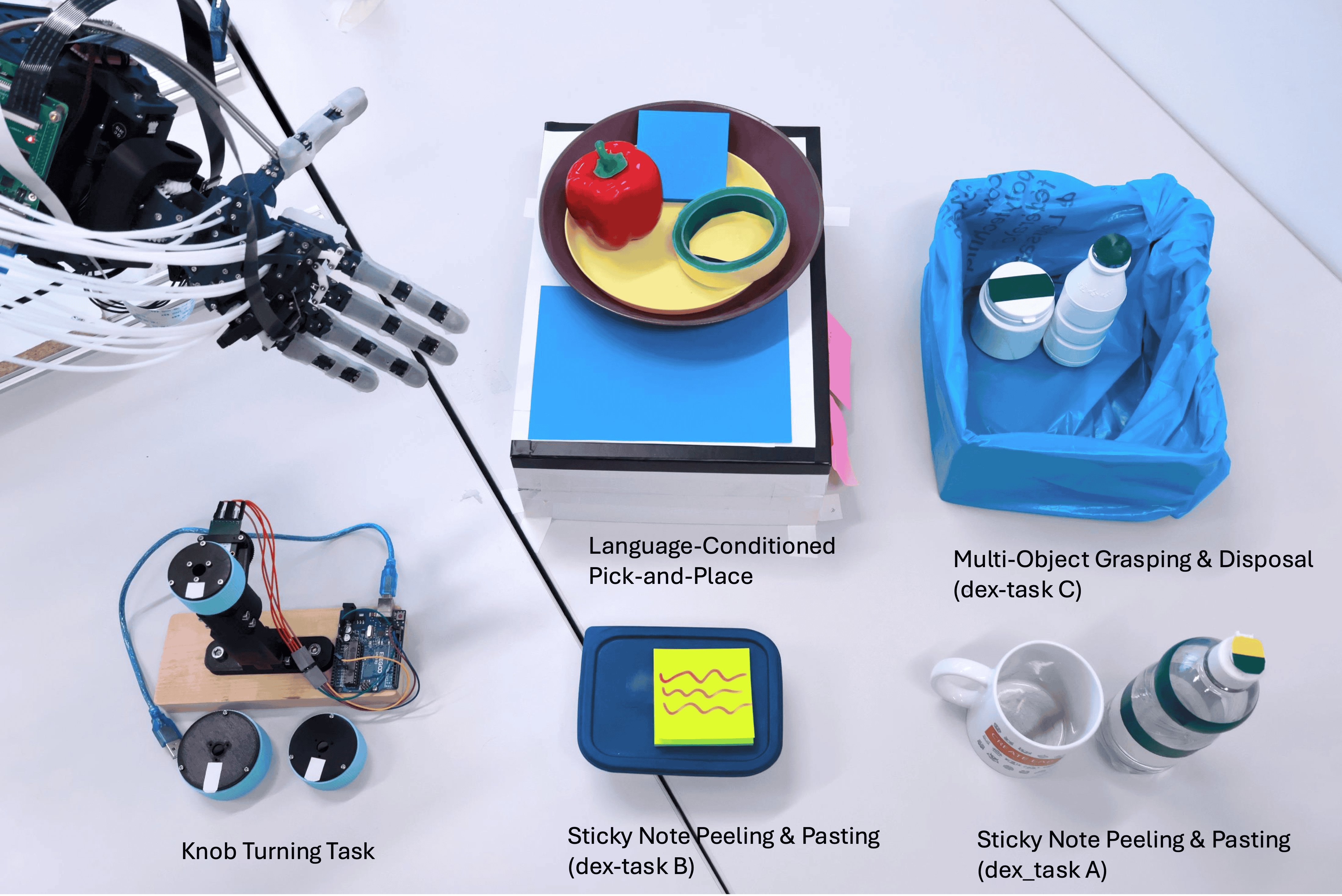}
     \caption{Overview of objects used in all experiments.}
     \label{si_fig:exp_obj}
\end{figure*}

\begin{figure*}[h]
     \centering
     \includegraphics[width=0.7\textwidth]{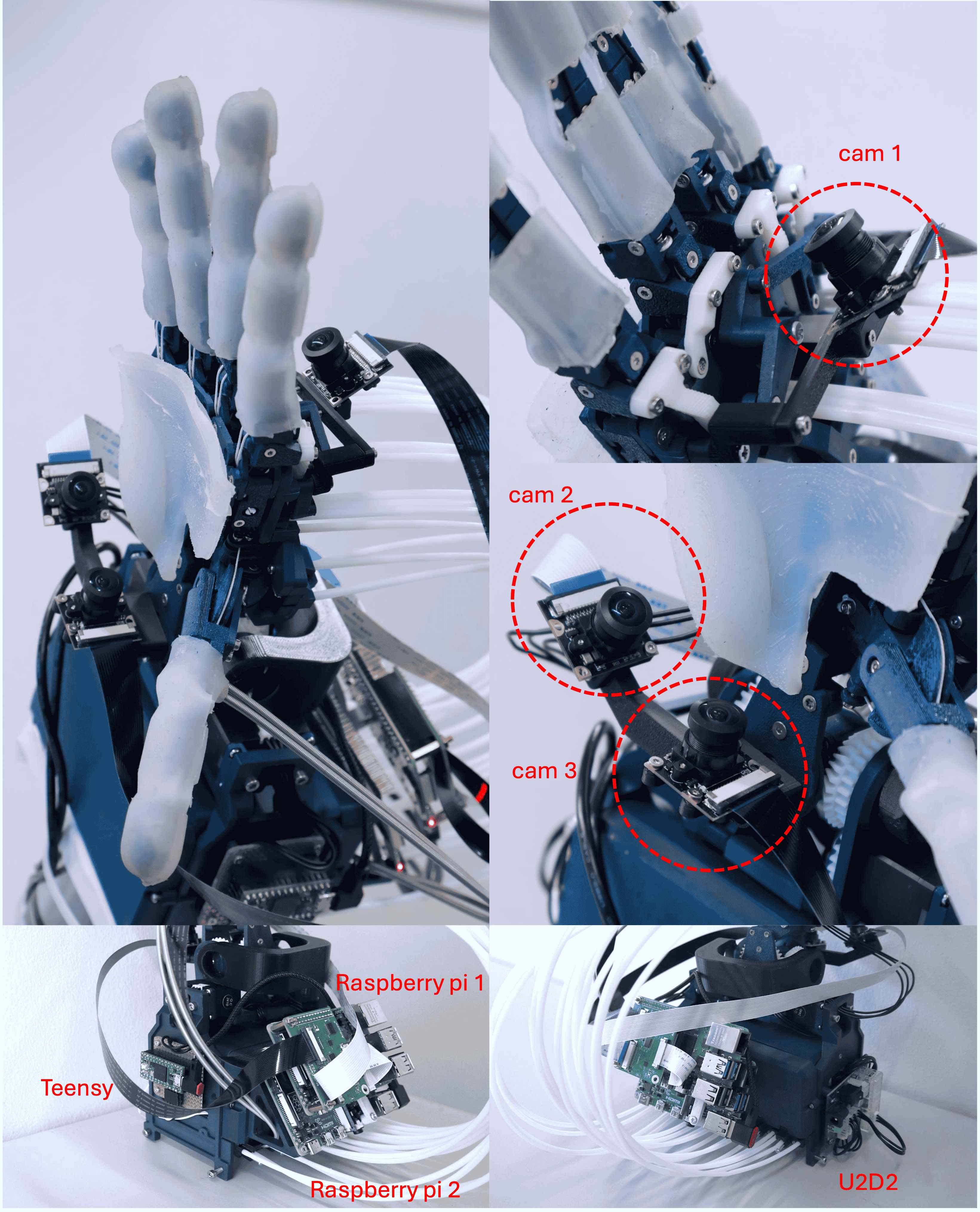}
     \caption{ADH robot hand, wrist cameras, and control or communication boards. Raspberry pi 1 (receiving and sending wrist camera images), Raspberry pi 2 (robot hand motion control, receiving and sending finger joints spring extensions), U2D2 (communication between Raspberry pi 2 and motors), Teensy (communication between Raspberry pi 2 and potentiometer positions for spring extension reading).}
     \label{si_fig:ADH}
\end{figure*}

\begin{figure*}[h]
     \centering
     \includegraphics[width=0.7\textwidth]{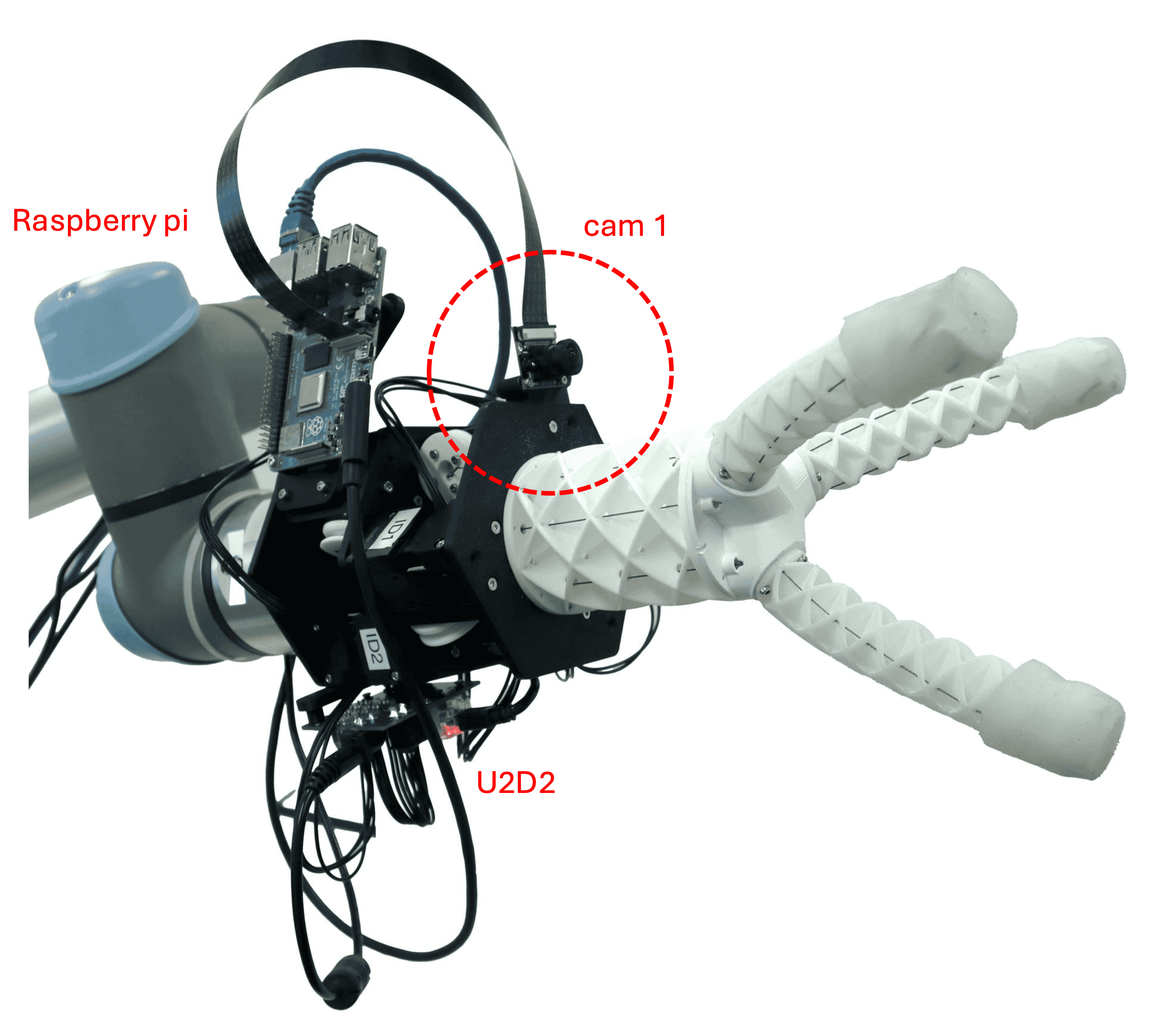}
     \caption{3-soft finger gripper, wrist camera, and control or communication boards. Raspberry pi (receiving and sending wrist camera images, and finger motion control), U2D2 (communication between Raspberry pi and motors).}
     \label{si_fig:softfingers}
\end{figure*}





\end{document}